%% file: main.tex
\documentclass{article} 

\PassOptionsToPackage{sort,numbers}{natbib} 

\usepackage[final]{neurips_data_2023}

\usepackage[utf8]{inputenc} 
\usepackage[T1]{fontenc}    
\usepackage[colorlinks=true, linkcolor=red, urlcolor=blue, citecolor=blue]{hyperref}
\usepackage{url}            
\usepackage{booktabs}       
\usepackage{amsfonts}       
\usepackage{nicefrac}       
\usepackage{microtype}      
\usepackage{xcolor}         

\input{macros}

\title{When Do Neural Nets Outperform Boosted Trees on Tabular Data?}

\author{Duncan McElfresh%
\thanks{Email: \texttt{duncan@abacus.ai, crwhite@caltech.edu}.
Work done while SK and JV were at Abacus.AI.}
~$^{1,2}$, Sujay Khandagale$^{3}$, Jonathan Valverde$^{4}$, Vishak Prasad C$^{5}$, \\
\textbf{Benjamin Feuer$^{6}$, Chinmay Hegde$^{6}$, Ganesh Ramakrishnan$^{5}$, Micah Goldblum$^{6}$, Colin White$^{1,7}$}
    \vspace*{1mm} \\
    $^1$ Abacus.AI, $^2$ Stanford, $^3$ Pinterest, $^4$ University of Maryland, \\
    $^5$ IIT Bombay, $^6$ New York University, $^7$ Caltech
  }

\begin{document}

\maketitle

\input{abstract}

\input{introduction}
\input{related_work}

\input{analysis}

\input{tabzilla}

\input{conclusion}

\section*{Author Contributions}
DM, SK, JV, VPC, GR, MG, and CW contributed to the original paper, which was accepted to the NeurIPS Datasets and Benchmarks Track 2023. BF and CH then joined to significantly extend and improve the results in Section 2 and Appendix D.



\bibliography{main}

\clearpage

\appendix

\input{impact}

\input{appendix_documentation}

\input{appendix_related_work}

\input{appendix_analysis}

\end{document}

%% file: macros.tex
\usepackage{longtable}
\usepackage{adjustbox}
\usepackage{array}

\usepackage{comment}
\usepackage{amsmath,amsthm,amssymb}
\usepackage{algorithm}
\usepackage{algpseudocode}
\usepackage{multicol}
\usepackage{mathtools}
\usepackage{caption}
\usepackage{float}
\usepackage{graphicx}
\usepackage{subcaption}
\usepackage{wrapfig}
\usepackage{multirow}
\usepackage{mathrsfs}
\usepackage{newfloat}
\usepackage{relsize}
\usepackage{enumitem}
\usepackage{pifont}
\usepackage{bm}

\usepackage{lscape}  

\usepackage{rotating}

\usepackage[nameinlink,capitalise,noabbrev]{cleveref}

\usepackage{xcolor}
\definecolor{dark-blue}{rgb}{0.15,0.15,0.4}

\hypersetup{
    colorlinks=true,
    linkcolor=blue,
    filecolor=magenta,      
    urlcolor=cyan,
    citecolor=dark-blue,
    pdftitle={When Do Neural Nets Outperform Boosted Trees on Tabular Data?},
    pdfpagemode=FullScreen,
}

\newcolumntype{R}[2]{%
    >{\adjustbox{angle=#1,lap=\width-(#2)}\bgroup}%
    l%
    <{\egroup}%
}

\newcolumntype{T}{>$l<$}
\newcolumntype{L}{>{\arraybackslash}m{4cm}}
\newcolumntype{S}{>{\arraybackslash}m{3cm}}

\usepackage[symbol]{footmisc}

\usepackage{tablefootnote}

\newcommand{\datasplit}{dataset split}

\newcommand{\dll}{$\Delta_{\ell\ell}$}%

\newcommand{\nalgs}{19}%

\newcommand{\ndatasets}{176}%
\newcommand{\nparams}{30}%
\newcommand{\nmodels}{538\,650}%

\newcommand{\nsuite}{36}%

%% file: abstract.tex
\begin{abstract}
Tabular data is one of the most commonly used types of data in machine learning. Despite recent advances in neural nets (NNs) for tabular data, there is still an active discussion on whether or not NNs generally outperform gradient-boosted decision trees (GBDTs) on tabular data, with several recent works arguing either that GBDTs consistently outperform NNs on tabular data, or vice versa.
In this work, we take a step back and question the importance of this debate.
To this end, we conduct the largest tabular data analysis to date, comparing \nalgs{} algorithms across \ndatasets{} datasets, and we find that the `NN vs.\ GBDT' debate is overemphasized: for a surprisingly high number of datasets, either the performance difference between GBDTs and NNs is negligible, or light hyperparameter tuning on a GBDT is more important than choosing between NNs and GBDTs.
A remarkable exception is the recently-proposed prior-data fitted network, TabPFN: although it is effectively limited to training sets of size 3000, we find that it outperforms all other algorithms on average, even when randomly sampling 3000 training datapoints.
Next, we analyze dozens of metafeatures to determine what \emph{properties} of a dataset make NNs or GBDTs better-suited to perform well. 
For example, we find that GBDTs are much better than NNs at handling skewed or heavy-tailed feature distributions and other forms of dataset irregularities.
Our insights act as a guide for practitioners to determine which techniques may work best on their dataset.
Finally, with the goal of accelerating tabular data research, we release the TabZilla Benchmark Suite: a collection of the \nsuite{} `hardest' of the datasets we study. 
Our benchmark suite, codebase, and all raw results are available at \url{https://github.com/naszilla/tabzilla}.
\end{abstract}

%% file: introduction.tex
\section{Introduction} \label{sec:introduction} 
Tabular datasets are data organized into rows and columns, consisting of distinct features that are typically continuous, categorical, or ordinal.
They are the oldest and among the most ubiquitous dataset types in machine learning in practice \citep{shwartz2022tabular,borisov2021deep}, due to their numerous applications across
medicine \citep{johnson2016mimic,ulmer2020trust},
finance \citep{arun2016loan,clements2020sequential},
online advertising \citep{richardson2007predicting,mcmahan2013ad,guo2017deepfm},
and many other areas \citep{chandola2009anomaly,buczak2015survey,urban2021deep}.

Despite recent advances in neural network (NN) architectures for tabular data \citep{arik2021tabnet,Popov2020Neural}, there is still an active debate over whether or not NNs generally outperform gradient-boosted decision trees (GBDTs) on tabular data, with multiple works arguing either for \citep{kadra2021well,arik2021tabnet, Popov2020Neural,rubachev2022revisiting, levin2023transfer}
or against \citep{shwartz2022tabular,borisov2021deep,gorishniy2021revisiting,grinsztajn2022tree} 
NNs.
This is in stark contrast to other areas such as computer vision and natural language understanding, in which NNs have far outpaced competing methods \citep{alexnet,vaswani2017attention,gpt3,dosovitskiy2021image,bert}.

Nearly all prior studies of tabular data use fewer than 50 datasets or do not properly tune baselines \citep{tunguz2022trouble,lipton2019troubling}, putting the generalizability of these findings into question.
Furthermore, the bottom line of many prior works is to answer the question, `which performs better, NNs or GBDTs, in terms of the average rank across datasets' without searching for more fine-grained insights.

In this work, we take a completely different approach by focusing on the following points.
First, we question the importance of the `NN vs.\ GBDT' debate, by investigating the significance of algorithm selection.
Second, we analyze what \emph{properties} of a dataset make NNs or GBDTs better-suited to perform well.
We take a data-driven approach to answering these questions, conducting the largest tabular data analysis to date, by comparing \textbf{\nalgs{} algorithms each with up to \nparams{} hyperparameter settings, across \ndatasets{} datasets}, including datasets from the OpenML-CC18 suite \cite{bischl2017openml} and the OpenML Benchmarking Suite \cite{gijsbers2019open}. To assess performance differences across datasets, we consider dozens of metafeatures.
We use 10 folds for each dataset to further reduce the uncertainty of our results.

We find that
for a surprisingly high fraction of datasets, either a very simple baseline performs on par with the top algorithms; 
furthermore, for roughly one-third of all datasets, light hyperparameter tuning on CatBoost or ResNet increases performance more than choosing among GBDTs and NNs. 
These results show that for many tabular datasets, it is not necessary to try out many different NNs and GBDTs: \textbf{in many cases, a strong baseline or a well-tuned GBDT will suffice}.

We do find that over all \ndatasets{} datasets, the two best-performing algorithms are CatBoost \citep{prokhorenkova2018catboost} and TabPFN \citep{hollmann2022tabpfn}. The latter's performance is particularly interesting because TabPFN is a recently-proposed prior-data fitted network that can perform training and inference on small datasets in less than one second, however its runtime and memory usage scale quadratically with the number of training samples. Surprisingly, we show that even for large datasets, randomly selecting 3000 training datapoints is sufficient for TabPFN to achieve very strong performance.

Next, we run analyses to discover what \emph{properties} of datasets explain which methods, or families of methods, do or do not succeed.
We compute the correlations of various metafeatures with algorithm performance, and we demonstrate that these correlations are predictive.
Our main findings are as follows (also see \cref{fig:irregularity-scatter}): \textbf{dataset \emph{regularity} is predictive of NNs outperforming GBDTs} (for example, feature distributions that are less skewed and less heavy-tailed).
Furthermore, GBDTs tend to perform better on larger datasets.

Finally, with the goal of accelerating tabular data research, we release the \textbf{TabZilla Benchmark Suite}: a collection of the `hardest' of the \ndatasets{} datasets we studied.
We select datasets on which a simple baseline does not win, as well as datasets such that most algorithms do not reach top performance.

Our work provides a large set of tools for researchers and practitioners working on tabular data.
We provide the largest open-source codebase of algorithms and datasets in one interface, together with a set of the `hardest' datasets, and raw results (over 500K trained models) at \url{https://github.com/naszilla/tabzilla} for researchers and practitioners to more easily compare methods.
Finally, our metafeature insights can be used by researchers, to uncover the failure modes of tabular algorithms, and by practitioners, to help determine which algorithms will perform well on a new dataset.

\begin{figure}
    \centering
    \includegraphics[width=0.98\textwidth]{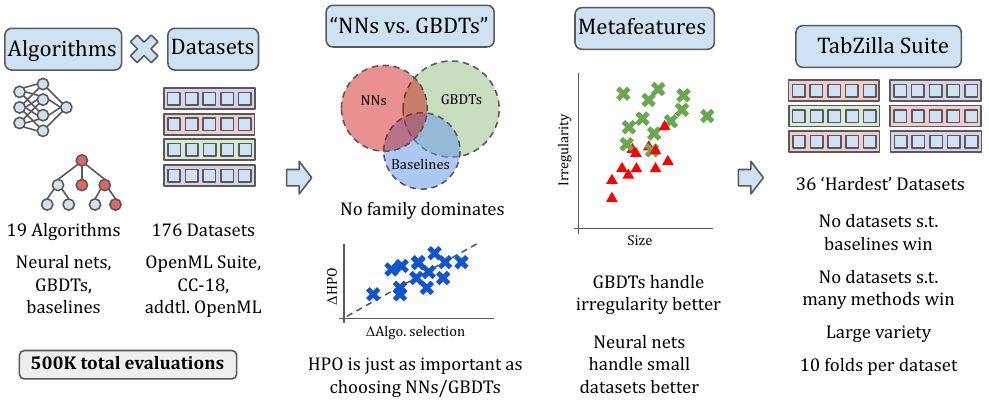}
    \caption{Overview of our work.
    We start by conducting the largest study on tabular data to date (left); we analyze the importance of algorithm selection (`NNs vs.\ GBDTs') as well as metafeatures (middle); and based on our study, we release TabZilla, a collection of the hardest tabular datasets.
    }
    \label{fig:tabzilla_overview}
\end{figure}

\noindent\textbf{Our contributions.}
We summarize our main contributions below:
\begin{itemize}[topsep=2pt, itemsep=2pt, parsep=0pt, leftmargin=5mm]
\item We conduct the largest analysis of tabular data to date, comparing \nalgs{} methods on \ndatasets{} datasets, 
with more than half a million models trained.
We show that for a surprisingly high fraction of datasets, either a simple baseline performs the best, or light hyperparameter tuning of a GBDT is more important than choosing among NNs and GBDTs, suggesting that the `NN vs.\ GBDT' debate is overemphasized.
\item After analyzing dozens of metafeatures, we present a number of insights into the properties that make a dataset better-suited for GBDTs or NNs.  
\item 
We release the TabZilla Suite: a collection of \nsuite{} `hard' datasets, with the goal of accelerating tabular data research.
We open-source our benchmark suite, codebase, and raw results.
\end{itemize}

%% file: related_work.tex
\paragraph{Related work.} 

Tabular datasets are the oldest and among the most common dataset types in machine learning in practice \citep{shwartz2022tabular,borisov2021deep}, due to their wide variety of applications \citep{johnson2016mimic,clements2020sequential,arun2016loan,richardson2007predicting,chandola2009anomaly}.
GBDTs \citep{friedman2001greedy} iteratively build an ensemble of decision trees, with each new tree fitting the residual of the loss from the previous trees, using gradient descent to minimize the losses.
XGBoost \citep{chen2016xgboost}, LightGBM \citep{ke2017lightgbm}, and Catboost \citep{prokhorenkova2018catboost} are three widely-used, high-performing variants.
Borisov et al.\ described three types of tabular data approaches for neural networks \citep{borisov2021deep}:
data transformation methods \citep{yoon2020vime,hancock2020survey}, architecture-based methods \citep{Popov2020Neural,guo2017deepfm,chen2022danets,arik2021tabnet} (including transformers \citep{gorishniy2021revisiting,huang2020tabtransformer,somepalli2021saint}), and regularization-based methods \citep{shavitt2018regularization,kadra2021well,jeffares2022tangos}.
Several recent works compare GBDTs to NNs on tabular data, often finding that \emph{either} NNs \citep{kadra2021well,gorishniy2021revisiting,arik2021tabnet,Popov2020Neural}
\emph{or} GBDTs \citep{shwartz2022tabular,borisov2021deep,grinsztajn2022tree,gorishniy2021revisiting}
perform best.

Perhaps the  most related work to ours is by Grinsztajn et al.\ \citep{grinsztajn2022tree}, who investigate why tree-based methods outperform neural nets on tabular data.
There are a few differences between their work and ours.
First, they consider seven algorithms and 45 datasets, compared to our 19 algorithms and 176 datasets.
Second, their dataset sizes range from 3\,000 to 10\,000, or seven that are exactly 50\,000, in contrast to our dataset sizes which range from 32 to 1\,025\,009 (see Table \ref{tab:dataset-stats}).
Additionally, they further control their study, for example by upper bounding the ratio of size to features, by removing high-cardinality categorical feature, and by removing low-cardinality numerical features. While this has the benefit of being a more controlled study, their analysis misses out on some of our observations, such as GBDTs performing better than NNs on `irregular' datasets.
Finally, while Grinsztajn et al.\ focused in depth on a few metafeatures such as dataset smoothness and number of uninformative features, our work considers orders of magnitude more metafeatures. Again, while each approach has its own strengths, our work is able to discover more potential insights, correlations, and takeaways for practitioners.
To the best of our knowledge, the only related work has considered more than 50 datasets is TabPFN \citep{hollmann2022tabpfn}, which considered 179 datasets which are size 2\,000 or smaller.
See \cref{app:relatedwork} for a longer discussion of related work.

%% file: analysis.tex
\section{Analysis of Algorithms for Tabular Data}  \label{sec:analysis}

In this section, we present a large-scale study of techniques for tabular data across a wide variety of datasets.
Our analysis seeks to answer the following two questions.

\begin{enumerate}[topsep=1pt, itemsep=2pt, parsep=0pt, leftmargin=5mm]
\item How do algorithms (and algorithm families) compare across a wide variety of datasets?
\item 
What properties of a dataset are associated with algorithms (and families) outperforming others?
\end{enumerate}

\paragraph{Algorithms and datasets implemented.}
We present results for \nalgs{} algorithms, including popular recent techniques and baselines.
The methods include three GBDTs: 
CatBoost \citep{prokhorenkova2018catboost}, 
LightGBM \citep{ke2017lightgbm}, and
XGBoost \citep{chen2016xgboost};
11 neural networks:
DANet \citep{chen2022danets},
FT-Transformer \citep{gorishniy2021revisiting},
two MLPs \citep{gorishniy2021revisiting},
NODE \citep{Popov2020Neural},
ResNet \citep{gorishniy2021revisiting},
SAINT \citep{somepalli2021saint},
STG \citep{yamada2020feature},
TabNet \citep{arik2021tabnet},
TabPFN \citep{hollmann2022tabpfn}, and
VIME \citep{yoon2020vime};
and five baselines:
Decision Tree \citep{quinlan1986induction},
KNN \citep{cover1967nearest},
Logistic Regression \citep{cox1958regression},
Random Forest \citep{liaw2002classification}, and
SVM \citep{cortes1995support}.
We choose these algorithms because of their popularity, diversity, and strong performance.
For TabPFN, the runtime and memory usage scale quadratically with the number of inputs (i.e., training samples). 
In order to run on datasets of size larger than 3000, we simply take a random sample of size 3000 from the full training dataset.
We denote this variant as TabPFN$^*$.

We run the algorithms on \ndatasets{} classification datasets from OpenML \citep{vanschoren2014openml}.
Our aim is to include most classification datasets from popular recent papers that study tabular data \citep{kadra2021well,borisov2021deep,shwartz2022tabular,gorishniy2021revisiting}, including datasets from the OpenML-CC18 suite \cite{bischl2017openml}, the OpenML Benchmarking Suite \cite{gijsbers2019open}, and additional OpenML datasets.
Due to the scale of our experiments (\nmodels{} total models trained), we limit the run-time for each experiment (described below), which precluded the use of datasets of size larger than 1.1M.
\cref{tab:dataset-stats} shows summary statistics for all datasets.
CC-18 and OpenML Benchmarking Suite are both seen as the go-to standards for conducting a fair, diverse evaluation across algorithms due to their rigorous selection criteria and wide diversity of datasets \citep{bischl2017openml,gijsbers2019open}.
To the best of our knowledge, our \nalgs{} algorithms and \ndatasets{} datasets are the largest number of \emph{either} algorithms \emph{or} datasets (with the exception of TabPFN \citep{hollmann2022tabpfn}) considered by recent tabular dataset literature, and the largest number available in a single open-source repository.

\paragraph{Metafeatures.}
We extract metafeatures using the Python library PyMFE \cite{pymfe}, which contains 965 metafeatures. 
The categories of metafeatures include: `general' (such as number of datapoints, classes, or numeric/categorical features), `statistical' (such as the min, mean, or max skewness, or kurtosis, of all feature distributions), `information theoretic' (such as the Shannon entropy of the target), `landmarking' (the performance of a baseline such as 1-Nearest Neighbor on a subsample of the dataset), and `model-based' (summary statistics for some model fit on the data, such as number of leaf nodes in a decision tree model). 
Since some of these features have long-tailed distributions, we also include the log of each strictly-positive metafeature in our analysis.

\begin{table}[t]
\centering
\caption{Performance of algorithms across 98 datasets (see \cref{tab:ranks-accuracy-all-datasets-all-algs} for extended results). Columns show the algorithm family (GBDT, NN, PFN, or baseline), rank over all datasets, the average normalized accuracy (Mean Acc.), the std.\ dev.\ of normalized accuracy across folds (Std.\ Acc.), and the train time in seconds per 1000 instances. Min/max/mean/median of these quantities are taken over all datasets.}
\label{tab:main-performance}
\resizebox{\linewidth}{!}{
\begin{tabular}{lcrrrrrrrrrr}
\toprule
{} & {} & \multicolumn{4}{l}{Rank} & \multicolumn{2}{l}{Mean Acc.} & \multicolumn{2}{l}{Std. Acc.} & \multicolumn{2}{l}{Time /1000 inst.} \\
Algorithm & Class &               min & max &   mean & med.\ &                           mean & med.\ &                          mean & med.\ &                              mean &  med.\ \\
\midrule
CatBoost & GBDT & 1 & 18 & 5.50 & 4 & 0.87 & 0.93 & 0.30 & 0.22 & 21.70 & 2.08 \\
TabPFN$^*$ & PFN & 1 & 18 & 5.89 & 5 & 0.83 & 0.92 & 0.27 & 0.19 & 0.25 & 0.01 \\
XGBoost & GBDT & 1 & 17 & 6.87 & 6 & 0.81 & 0.89 & 0.33 & 0.22 & 0.81 & 0.37 \\
ResNet & NN & 1 & 19 & 7.65 & 8 & 0.75 & 0.83 & 0.30 & 0.21 & 16.01 & 9.34 \\
SAINT & NN & 1 & 19 & 7.90 & 7 & 0.73 & 0.86 & 0.31 & 0.24 & 169.54 & 146.16 \\
NODE & NN & 1 & 19 & 7.90 & 7 & 0.74 & 0.81 & 0.26 & 0.20 & 138.36 & 117.04 \\
FTTransformer & NN & 1 & 18 & 8.14 & 8 & 0.76 & 0.80 & 0.31 & 0.21 & 27.67 & 18.40 \\
RandomForest & base & 1 & 19 & 8.26 & 7 & 0.76 & 0.83 & 0.32 & 0.22 & 0.35 & 0.24 \\
LightGBM & GBDT & 1 & 19 & 8.46 & 8 & 0.76 & 0.84 & 0.36 & 0.21 & 0.87 & 0.34 \\
SVM & base & 1 & 18 & 9.09 & 10 & 0.69 & 0.76 & 0.26 & 0.19 & 30.40 & 1.67 \\
DANet & NN & 1 & 18 & 9.73 & 10 & 0.73 & 0.79 & 0.32 & 0.23 & 68.82 & 60.15 \\
    MLP-rtdl & NN & 1 & 19 & 9.85 & 10.50 & 0.65 & 0.72 & 0.28 & 0.16 & 14.27 & 7.30 \\
STG & NN & 1 & 19 & 11.76 & 12.50 & 0.56 & 0.63 & 0.29 & 0.17 & 18.44 & 15.79 \\
DecisionTree & base & 1 & 19 & 11.81 & 13 & 0.59 & 0.68 & 0.35 & 0.25 & 0.03 & 0.01 \\
MLP & NN & 1 & 19 & 12.00 & 13 & 0.57 & 0.57 & 0.29 & 0.18 & 18.39 & 11.20 \\
LinearModel & base & 1 & 19 & 12.18 & 14 & 0.51 & 0.53 & 0.31 & 0.24 & 0.04 & 0.03 \\
TabNet & NN & 1 & 19 & 12.69 & 14 & 0.54 & 0.60 & 0.39 & 0.25 & 34.95 & 29.90 \\
KNN & base & 1 & 19 & 13.69 & 15 & 0.45 & 0.51 & 0.29 & 0.21 & 0.01 & 0.00 \\
VIME & NN & 2 & 19 & 14.98 & 17 & 0.37 & 0.32 & 0.27 & 0.18 & 16.81 & 14.86 \\
\bottomrule
\end{tabular}
}
\end{table}

\paragraph{Experimental design.}
For each dataset, we use the ten train/test folds provided by OpenML, which allows our results on the test folds to be compared with other works that used the same OpenML datasets.
Since we also need validation splits in order to run hyperparameter tuning, we divide each training fold into a training and validation set. 
%
For each algorithm, and for each \datasplit{}, we run the algorithm for up to 10 hours. During this time, we train and evaluate the algorithm with at most 30 hyperparameter sets (one default set and 29 random sets, using Optuna \citep{akiba2019optuna}). 
Each parameterized algorithm is given at most two hours on a 32GiB V100 to complete a single train/evaluation cycle.
In line with prior work, our main metric of interest is \emph{accuracy}, and we report the test performance of the hyperparameter setting that had the maximum performance on the validation set.
We also consider log loss, which is highly correlated with accuracy but contains significantly fewer ties. We also include results for F1-score and ROC AUC in \cref{app:experiments}. 
Similar to prior work \citep{wistuba2015learning,feurer2022auto}, whenever we average across datasets, we use the average distance to the minimum (ADTM) metric, which consists of 0-1 scaling (after selecting the best hyperparameters, which helps protect against outliers \citep{grinsztajn2022tree}).
Finally, in order to see the variance of each method on different folds of the same dataset, we report the average (scaled) standard deviation of each method across all 10 folds.

\begin{table}
    \centering
{\small
\caption{Performance of algorithms across 57 datasets of size less than or equal to 1250. Columns show the rank over all datasets, the average normalized accuracy (Mean Acc.), the standard deviation of normalized accuracy across folds (Std.\ Acc), and the train time per 1000 instances. Min/max/mean/median are taken over all datasets.}
\label{tab:tabpfn-ranks-acc}
\begin{tabular}{lrrrrrrrrrr}
\toprule
{} & \multicolumn{4}{l}{Rank} & \multicolumn{2}{l}{Mean Acc.} & \multicolumn{2}{l}{Std.\ Acc.} & \multicolumn{2}{l}{Time /1000 inst.} \\
Algorithm &                min & max &   mean & med.\ &                           mean & med.\ &                          mean & med.\ &                              mean &  med.\ \\
\midrule
TabPFN$^*$        &                  1 &  18 &   4.88 &      3 &                           0.84 &   0.93 &                          0.35 &   0.26 &                              0.00 &    0.00 \\
CatBoost      &                  1 &  18 &   5.37 &      4 &                           0.85 &   0.91 &                          0.39 &   0.30 &                             26.22 &    2.75 \\
ResNet        &                  1 &  19 &   6.75 &      6 &                           0.77 &   0.79 &                          0.42 &   0.30 &                             23.67 &   13.87 \\
RandomForest  &                  1 &  18 &   7.65 &      7 &                           0.76 &   0.82 &                          0.40 &   0.29 &                              0.47 &    0.32 \\
SAINT         &                  1 &  19 &   7.67 &      6 &                           0.74 &   0.87 &                          0.42 &   0.31 &                            197.41 &  181.62 \\
FTTransformer &                  1 &  18 &   7.93 &      7 &                           0.75 &   0.78 &                          0.42 &   0.32 &                             32.93 &   26.39 \\
XGBoost       &                  1 &  17 &   8.30 &      8 &                           0.74 &   0.80 &                          0.42 &   0.30 &                              0.95 &    0.61 \\
NODE          &                  1 &  19 &   8.35 &      8 &                           0.73 &   0.75 &                          0.36 &   0.28 &                            173.55 &  144.45 \\
SVM           &                  1 &  18 &   9.54 &     11 &                           0.68 &   0.72 &                          0.35 &   0.28 &                             23.90 &    0.42 \\
MLP-rtdl      &                  1 &  19 &   9.77 &     10 &                           0.64 &   0.69 &                          0.39 &   0.31 &                             21.48 &   12.21 \\
LightGBM      &                  1 &  19 &  10.00 &     10 &                           0.68 &   0.71 &                          0.45 &   0.38 &                              0.64 &    0.23 \\
LinearModel   &                  1 &  19 &  10.21 &     11 &                           0.61 &   0.71 &                          0.38 &   0.29 &                              0.06 &    0.05 \\
DANet         &                  1 &  18 &  10.74 &     10 &                           0.68 &   0.69 &                          0.41 &   0.34 &                             83.57 &   71.19 \\
DecisionTree  &                  1 &  19 &  11.44 &     13 &                           0.60 &   0.67 &                          0.45 &   0.32 &                              0.02 &    0.01 \\
MLP           &                  1 &  19 &  11.49 &     13 &                           0.57 &   0.54 &                          0.38 &   0.30 &                             27.88 &   16.81 \\
STG           &                  1 &  19 &  11.49 &     12 &                           0.57 &   0.64 &                          0.40 &   0.34 &                             21.22 &   18.24 \\
KNN           &                  1 &  19 &  13.12 &     15 &                           0.46 &   0.51 &                          0.38 &   0.32 &                              0.00 &    0.00 \\
TabNet        &                  3 &  19 &  14.54 &     16 &                           0.42 &   0.40 &                          0.52 &   0.49 &                             41.83 &   34.35 \\
VIME          &                  2 &  19 &  14.88 &     17 &                           0.33 &   0.27 &                          0.36 &   0.29 &                             18.95 &   16.43 \\
\bottomrule
\end{tabular}
}
\end{table}


\subsection{Relative Algorithm Performance} \label{subsec:relative-performance}

In this section, we answer the question, ``How do individual algorithms, and families of algorithms, perform across a wide variety of datasets?''
We especially consider whether the difference between GBDTs and NNs is significant.

\paragraph{No individual algorithm dominates.}
We start by comparing the average rank of all algorithms across all datasets, while excluding datasets which ran into memory or timeout issues on a nontrivial number of algorithms
using the experimental setup described above. 
Therefore, we consider a set of 98 datasets (and we include results on all \ndatasets{} datasets in the next section and in \cref{app:relative-performance}).
As mentioned in the previous section, for each algorithm and dataset split, we report the test set performance after tuning on the validation set; see \cref{tab:main-performance}.

Surprisingly, \emph{nearly every algorithm ranks first on at least one dataset and last on at least one other dataset}.
As expected, baseline methods tend to perform poorly while neural nets and GBDTs tend to perform better on average. 
The fact that the best out of all algorithms, CatBoost, only achieved an average rank of 5.06, shows that there is not a single approach that dominates across most datasets.

\paragraph{TabPFN performs remarkably well.}
In \cref{tab:main-performance}, we find that TabPFN achieves nearly the same performance as CatBoost.
This is particularly surprising for two reasons. First, TabPFN's average training time per 1000 instances is 0.25 seconds: the fastest of any non-baseline, and two orders of magnitude faster than CatBoost.
Second, recall that in order to run TabPFN on large datasets, we simply run TabPFN with a random sample of size 3000 from the full training dataset. TabPFN achieves near-top performance, despite only seeing part of the training dataset for many datasets.

Next, in \cref{tab:tabpfn-ranks-acc}, we compute the same table for the 57 smallest datasets: the datasets with size at most 1250.
Now, \textbf{we find that TabPFN achieves the best average performance of all algorithms, while also having the fastest training time.} 
However, one caveat is that the \emph{inference} time for TabPFN is 2.36 seconds per 1000 instances, which is higher than other algorithms.
We give further discussion and results on TabPFN in \cref{app:tabpfn}, including an ablation study of randomly sampling either 1k or 3k training points, for both TabPFN and CatBoost.

\begin{wrapfigure}{r}{0.6\textwidth}
\vspace{-6mm}
\begin{center}
    \centering
\includegraphics[width=0.6\textwidth]{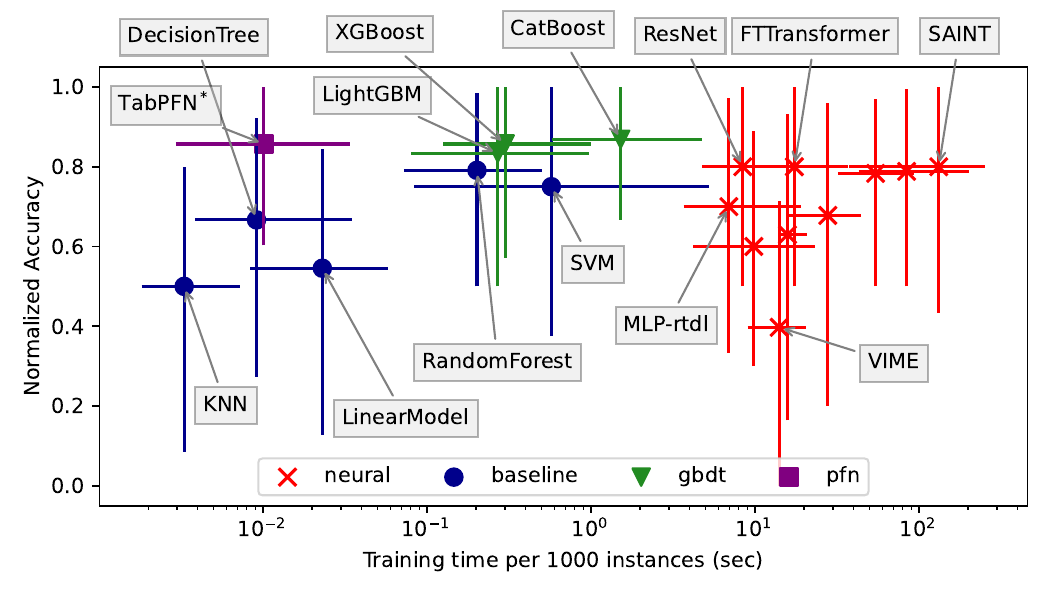}
\end{center}
\caption{Median runtime vs.\ median normalized accuracy for each algorithm, over 98 datasets. The bars span the 20th to 80th percentile over all datasets.}
\label{fig:runtime-vs-acc-annotations}
\vspace{-3mm}
\end{wrapfigure}

\paragraph{Performance vs.\ runtime.}
In \cref{fig:runtime-vs-acc-annotations}, we plot the accuracy vs.\ runtime for all algorithms, averaged across all datasets.
Overall, neural nets require the longest runtime, and often outperform baseline methods. 
On the other hand, GBDTs simultaneously require little runtime while also achieving strong performance: they consistently outperform baseline methods, and consistently require less runtime than NNs. 
Once again, we see the impressive performance of TabPFN, which achieves top accuracy with less training time than any GBDT or NN.

\begin{figure}
    \centering
    \includegraphics[width=0.8\textwidth]{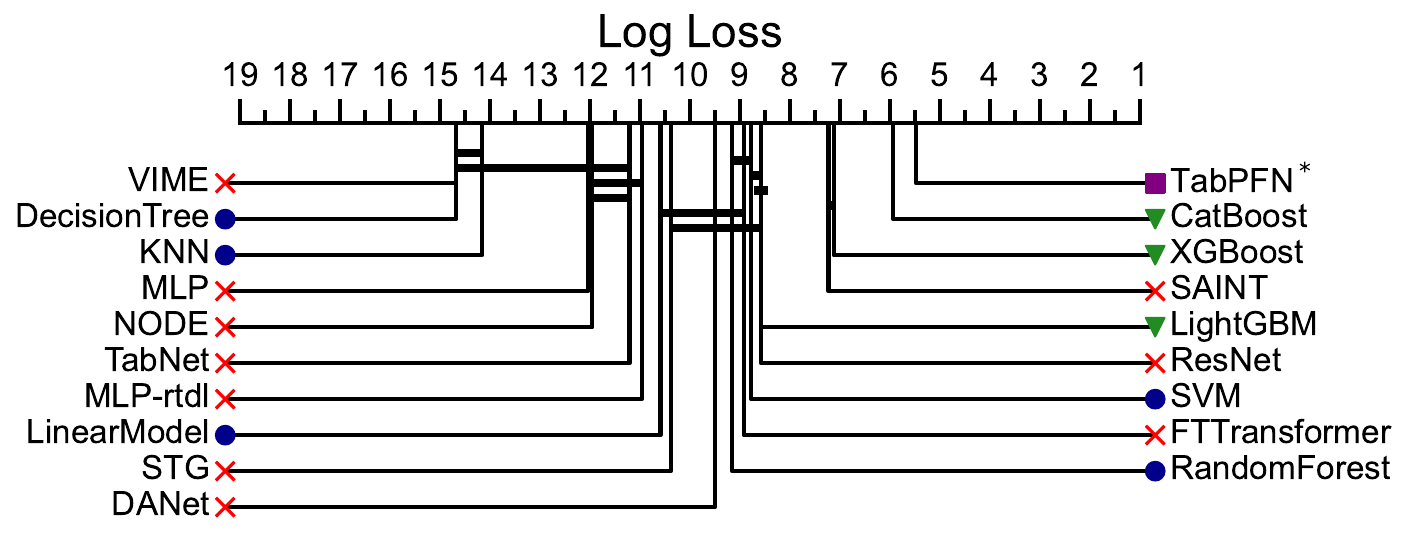}
    \caption{Critical difference plot comparing all algorithms according to their mean log loss rank over 98 datasets. Each algorithm's average rank is shown as a horizontal line on the axis. Sets of algorithms which are \emph{not significantly different} are connected by a horizontal black bar. Algorithm family is indicated by a marker next to the algorithm name: red ``X'' indicates a neural net, blue circle indicates a baseline algorithm, green triangles indicate GBDTs, and purple squares indicate a PFN.
    }
    \label{fig:cd-all-algs-ll}
\end{figure}

\paragraph{Statistically significant performance differences.}
\cref{tab:main-performance} shows that many algorithms have similar performance.
Next, we determine \emph{statistically significant} (p<$0.05)$ performance differences between algorithms across the 98 datasets described above.
First, we use a Friedman test to determine whether performance differences between each algorithm are significant \citep{friedman1937use}; we can reject the null hypothesis (p<$0.05$) for this test; the p-value is less than $10^{-20}$.
We then use a Wilcoxon signed-rank test to determine which pairs of algorithms have significant performance differences (p<$0.05$) \citep{conover1999practical}. 
With the Wilcoxon tests we use a Holm-Bonferroni correction to account for multiple comparisons \citep{holm1979simple}. 
Due to the presence of many ties in the test accuracy metric, we use the test log loss metric. See \cref{fig:cd-all-algs-ll} (and see \cref{fig:cd-all-algs-f1} for the F1 score).
In these figures, the average rank of each algorithm is shown on the horizontal axis; if differences between algorithms are \emph{not significant} (p$\geq0.05$), then algorithms are shown connected by a horizontal bar.
We find that TabPFN outperforms all other algorithms on average across 98 datasets, and this result is statistically significant.
Note that the slight differences between \cref{fig:cd-all-algs-ll} and \cref{tab:main-performance} is that the former uses accuracy, while the latter uses log loss.

\begin{figure}
    \centering
    \includegraphics[width=0.48\linewidth]{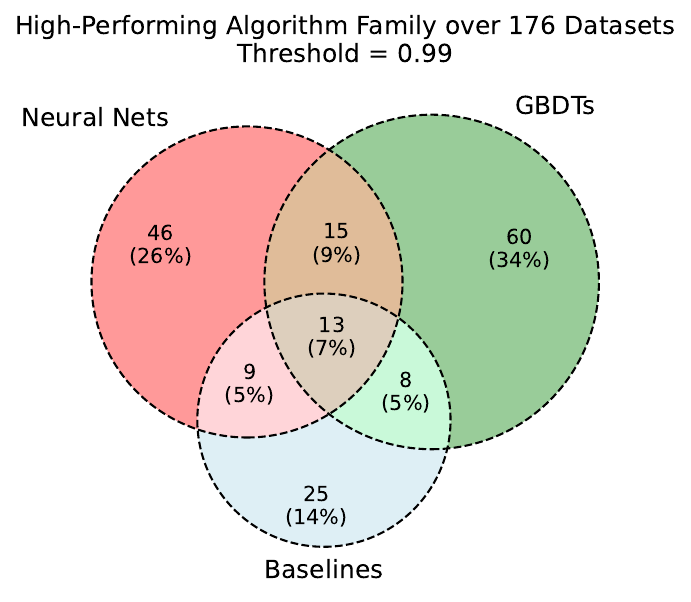}
    \includegraphics[width=0.45\textwidth]{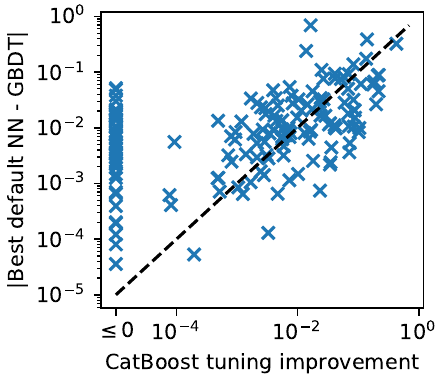}
    \caption{ 
    Left: Venn diagram of the number datasets where each algorithm is `high-performing' for each algorithm class, over all \ndatasets{} datasets. An algorithm is high-performing if its test accuracy after 0-1 scaling is at least 0.99 (we show 0.9999 in \cref{app:relative-performance}).
    Right: the performance improvement of hyperparameter tuning on CatBoost, compared to the absolute performance difference between the best neural net and the best GBDT using default hyperparameters. 
    Each point indicates the normalized log loss of one dataset,
    Points on or below the dotted line indicate that the performance improvement due to tuning is as high as the difference between NN-GBDT algorithm selection.}
    \label{fig:venn}
\end{figure}

\paragraph{GBDTs vs.\ NNs.}
Although \cref{tab:ranks-accuracy-all-datasets-all-algs} tells us which \emph{individual} methods perform best on average, 
now we consider the age-old question, `are GBDTs better than NNs for tabular data?'
We split the \nalgs{} algorithms into three \emph{families}: GBDTs (CatBoost, XGBoost, LightGBM), NNs (DANet, FT-Transformer, two MLPs, NODE, ResNet, SAINT, STG, TabNet, and VIME), and baselines (Decision Tree, KNN, LinearModel, RandomForest, SVM).
\footnote{
Since TabPFN works substantially differently from any neural net (or any other algorithm), we exclude it from our analysis in this section and the next section when discussing `GBDTs vs.\ NNs.'
}
We say that an algorithm is `high-performing' if it achieves a 0-1 scaled test accuracy of at least $0.99$, and then we determine which algorithm families (GBDTs, NNs, baselines) have a high-performing algorithm; see \cref{fig:venn}.
Surprisingly, the three-way Venn diagram is relatively balanced among GBDTs, NNs, and baselines, although GBDTs overall have the edge.
In \cref{app:relative-performance}, we run the same analysis, using a threshold of $0.9999$.
In this case, GBDTs are the sole high-performing algorithm family for most datasets.
Since these wins are by less than 0.01\%, they may not be significant to practitioners.

\paragraph{Algorithm selection vs.\ tuning.}
Next, we determine whether it is more important to select the best possible algorithm family, or to simply run light hyperparameter tuning on an algorithm that performs well in general, such as CatBoost or ResNet.
We consider a scenario in which a practitioner can decide to \emph{(a)} test several algorithms using their default hyperparameters, or \emph{(b)} optimize the hyperparameters of a single model, such as CatBoost or ResNet. 
We compute whether \emph{(a)} or \emph{(b)} leads to better performance.
Specifically, we measure the performance difference between the best-performing GBDT and NN using their default hyperparameters, as well as the performance difference between CatBoost with the default hyperparameters vs.\ CatBoost tuned via 30 iterations of random search on a validation set; see \cref{fig:venn} (right), and see \cref{app:relative-performance} for the same analysis with ResNet.
Surprisingly, light hyperparameter tuning yields a greater performance improvement than GBDT-vs-NN selection for about one-third of all datasets.
Once again, this suggests that for a large fraction of datasets, it is not necessary to determine whether GBDTs or NNs are better: light tuning on an algorithm such as CatBoost or ResNet can give just as much performance gain.
In the next section, we explore \emph{why} a dataset might be more amenable to a neural net or a GBDT.


\subsection{Metafeature Analysis} \label{subsec:metafeature}
In this section, we answer the question,
``What properties of a dataset are associated with certain techniques, or families of techniques, outperforming others?''
We answer this question by computing the correlation of metafeatures with three different quantities related to the performance difference between algorithm families, the performance difference between pairs of algorithms, and the relative performance of individual algorithms.

\begin{figure}
    \centering
    \includegraphics[width=0.48\textwidth]{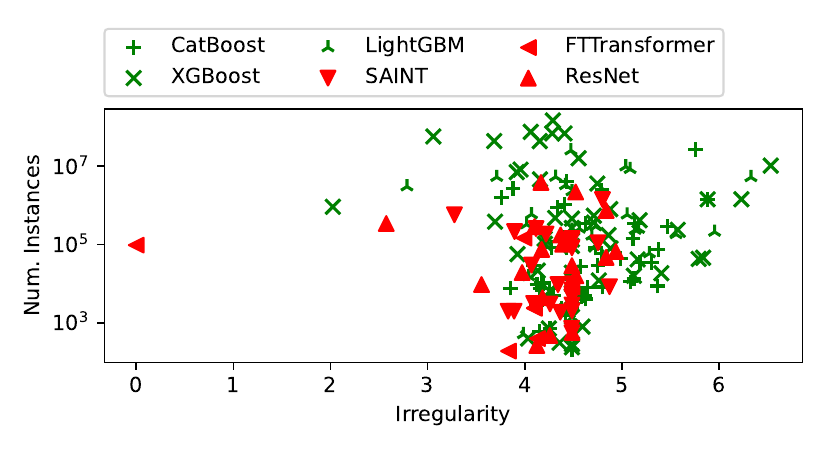}
    \includegraphics[width=0.48\textwidth]{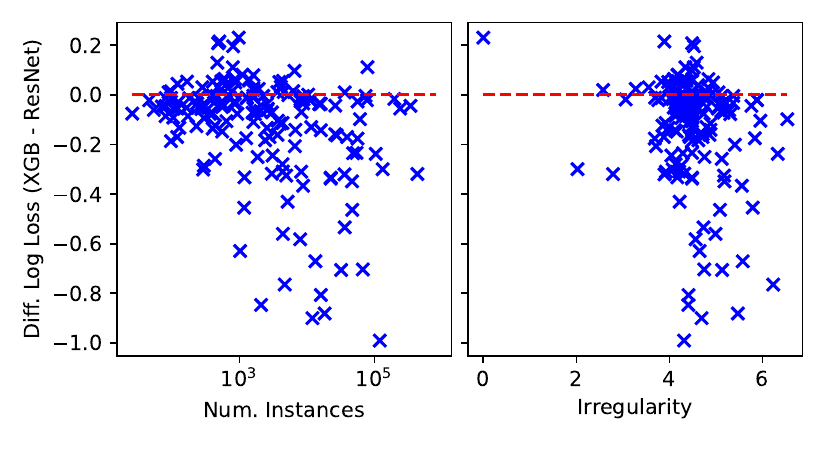}
    \caption{Left: scatterplot of the best algorithm on all \ndatasets{} datasets across metafeatures. The vertical axis indicates the dataset size, and the horizontal axis combines five dataset metafeatures related to irregularity. Right: scatterplot of the difference in normalized log loss between XGBoost and ResNet, by dataset size (middle subplot) and irregularity (right subplot). The irregularity feature is a linear combination of five standardized dataset attributes: the minimum eigenvalue of the feature covariance matrix (-0.33), the skewness of the standard deviation of all features (0.23), the skewness of the range of all features (0.22), the interquartile range of the harmonic mean of all features (0.21), and the standard deviation of the kurtosis of all features (0.21).}
    \label{fig:irregularity-scatter}
\end{figure}

In order to assess the difference in performance between NNs and GBDTs, we calculate the difference in normalized log loss between the best NN and the best GBDT, which we refer to as \dll{}. 
We compute the correlation of \dll{} to various metafeatures across all datasets; see \cref{fig:corr-nn-gbdt-diff-new} and \cref{tab:nn-gbdt-corr-new} in \cref{app:metafeature}.
Next, in order to determine the individual strengths and weaknesses of each algorithm, we compute the correlation of various metafeatures to the performance of an individual algorithm relative to all other algorithms; see
\cref{tab:cor-with-alg-performance} and \cref{fig:irregularity-scatter}.
Finally, we compute the correlation of metafeatures to the difference in performance between pairs of the top-performing algorithms from \cref{subsec:relative-performance}: CatBoost, XGBoost, SAINT, and ResNet.
See \cref{fig:irregularity-scatter} and \cref{tab:feature-corrs-ll}.

In order to show that these metafeatures are \emph{predictive}, we train and evaluate a meta-learning model using a leave-one-out approach: one dataset is held out for testing, while the remaining 175 datasets are used for training, averaged across all \ndatasets{} possible test sets; see \cref{app:metafeature}.
In the rest of this section, we state and discuss the main findings of our metafeature analysis.

\paragraph{Neural nets perform comparatively worse on larger datasets.}
Throughout our metafeature analyses, we find that GBDTs perform comparatively better than NNs and baselines with larger datasets.
\cref{fig:irregularity-scatter} shows that XGBoost achieves top performance compared to all \nalgs{} algorithms on the seven largest datasets, and GBDTs overall perform well.
In \cref{tab:cor-with-alg-performance}, somewhat surprisingly, dataset size is the most negatively-correlated metafeature with the relative performance of both LightGBM and XGBoost.
Finally, \cref{tab:nn-gbdt-corr-new} shows that the GBDT family's performance is also positively correlated with the \emph{ratio} of size to the number of features.
Notably, all of these analyses are relative to the performance of all algorithms, which includes the newly released TabPFN \citep{hollmann2022tabpfn}, a neural net that performs remarkably well on datasets of small size, due to its carefully-designed prior.
On the other hand, GBDTs excel when the ratio of dataset size to number of features is high, because all split in the decision trees are computed using more datapoints.
Some of our above findings are backed up by prior work, for example, Grinsztajn et al.\ \citep{grinsztajn2022tree} showed that increasing the ratio of (uninformative) features to dataset size, hurts the performance of ResNet, and our results indicate the same trend (\cref{tab:feature-corrs-ll-general}). On the other hand, NNs as a whole see the opposite trend (\cref{tab:nn-gbdt-corr-new}), which at least is shown for FTTransformer in Grinsztajn et al. \citep{grinsztajn2022tree}. 
We provide additional analyses in \cref{app:dataset-size}.
Note that, while the general trend shows GBDTs outperforming NNs on larger datasets, this does not imply that all GBDTs are better than all NNs on larger datasets. For example, TabPFN and TabNet are both neural nets, yet TabPFN performs particularly well on smaller datasets, and TabNet performs particularly well on larger datasets.
\textbf{It is important to note that when choosing an algorithm for a new use-case, practitioners should focus on algorithm-specific analyses (such as in \cref{tab:main-performance}, \cref{tab:tabpfn-ranks-acc}, and \cref{app:dataset-size}) rather than general `GBDT vs.\ NN' trends.}

\paragraph{GBDTs favor irregular datasets.}
Another trend present throughout all three metafeature analyses is that
GBDTs consistently favor `irregular' datasets.
When comparing pairs of the top GBDTs and NNs, we find that both CatBoost and XGBoost outperform ResNet and SAINT on datasets whose feature distributions are heavy-tailed, skewed, or have high variance (see \cref{tab:feature-corrs-ll} for the full details).
That is, some datasets' feature distributions all have a similar amount of skewness, while other datasets' feature distributions are more irregular, with a high range of skewness. It is the latter type of datasets on which GBDTs outperform NNs.
We also find that GBDTs perform better when datasets are more class imbalanced (on SAINT in particular).
In \cref{fig:irregularity-scatter}, GBDTs perform best on the most irregular datasets, computed via a linear combination of five metafeatures each measuring the skewness or kurtosis of the feature distributions.

\begin{table}[t]
\small
    \centering
    \caption{Metafeatures that are most correlated with the performance of each top-performing algorithm, calculated as the Pearson correlation between the metafeature and normalized log loss over 
    \ndatasets{} datasets. Metafeatures with the largest absolute Pearson correlation with algorithm performance are shown, and correlations are presented as 95\% confidence intervals, calculated using the Fisher transformation.}
    \label{tab:cor-with-alg-performance}
    \resizebox{\linewidth}{!}{
    \begin{tabular}{lp{11.5cm}l}\toprule
        Alg. & Metafeature Description & Corr. \\\midrule
        CatBoost & Noisiness of the features: $(\sum_i S_i - \sum_i MI(i, y)) / \sum_i MI(i, y)$, where $S_i$ is the entropy of feature $i$, and $MI(i, y)$ is the mutual information between feature $i$ and the target $y$.
         & [0.25, 0.34] \\\specialrule{0.2pt}{2pt}{2pt} 
         XGBoost & Log. number of instances. & [-0.27, -0.18] \\\specialrule{0.2pt}{2pt}{2pt} 
         LightGBM & Log. number of instances. & [-0.36, -0.28] \\\specialrule{0.2pt}{2pt}{2pt} 
         ResNet & Mean  canonical correlation between any numeric feature and the target. & [-0.28, -0.19] \\\specialrule{0.2pt}{2pt}{2pt} 
         FTTransformer & Number of target classes. &  [0.23, 0.32] \\\specialrule{0.2pt}{2pt}{2pt} 
         SAINT & Noisiness of the features. & [0.19, 0.29] \\\bottomrule 
    \end{tabular}
    }
\end{table}

\paragraph{The bottom line.}
Overall, we answer the title question of our paper: GBDTs outperform NNs on datasets that are more `irregular', 
as well as large datasets, and datasets with a high ratio of size to number of features.
When a practitioner is faced with a new dataset, based on all the analysis in \cref{sec:analysis}, we give the following recommendation: first try simple baselines, and then conduct light hyperparameter tuning on CatBoost.
Surprisingly, this often will already result in strong performance. 
As a next step, the practitioner can try NNs and other GBDTs that are most-correlated with strong performance based on the dataset's metafeatures, using analyses such as \cref{tab:main-performance} and \cref{app:dataset-size}.

%% file: tabzilla.tex
\section{TabZilla Benchmark Suite} \label{sec:tabzilla}

In order to accelerate tabular data research, we release the TabZilla Benchmark Suite: a collection of the \nsuite{} `hardest' of the \ndatasets{} datasets we studied in \cref{sec:analysis}.
We use the following three criteria.

\paragraph{Hard for baseline algorithms.} As discussed in \cref{subsec:relative-performance}, simple baselines perform very well on a surprisingly large fraction of the datasets in our experiments.
Therefore, to select our suite of hard datasets, we remove any dataset such that a baseline (as defined in \cref{sec:analysis}) achieved a normalized log loss within 20\% of the top-performing algorithm.
This criterion is not perfect; for example, if a dataset is so hard that all \nalgs{} algorithms fail to reach non-trivial performance, it would not satisfy the criterion. However, the criterion is a good proxy for dataset hardness given the available information.

\begin{table}[t]
\centering
\caption{\label{tab:benchmark-datasets} The TabZilla Benchmark Suite. 
Columns show the hardness metrics used as selection criteria, dataset attributes, and the top-performing algorithms.
`Std. Kurtosis' indicates the std.\ dev.\ of the kurtosis of all features. Hardness metrics that meet our selection criteria are shown in bold.
}
\resizebox{\linewidth}{!}{
\begin{tabular}{llllrrrlll}
\toprule
{} & \multicolumn{3}{c}{Hardness Metrics} & \multicolumn{3}{c}{Dataset Attributes} & \multicolumn{3}{c}{Top 3 Algs.}\\

                              Dataset &  base & 4th-best &  GBDT & $N$ &  \# feats. & Std. Kurtosis & 1st &           2nd &            3rd \\
\midrule
credit-g & \textbf{0.26} & \textbf{0.13} & \textbf{0.12} & 1 000 & 21 & 1.92 & ResNet & FTTransformer & CatBoost \\
jungle-chess & \textbf{0.30} & \textbf{0.18} & \textbf{0.17} & 44 819 & 7 & 0.08 & SAINT & TabNet & LightGBM \\
MiniBooNE & \textbf{0.20} & \textbf{0.09} & 0.00 & 130 064 & 51 & 12162.65 & LightGBM & XGBoost & CatBoost \\
albert & \textbf{0.42} & \textbf{0.28} & 0.00 & 425 240 & 79 & 1686.90 & CatBoost & XGBoost & ResNet \\
electricity & \textbf{0.46} & \textbf{0.38} & 0.00 & 45 312 & 9 & 2693.51 & LightGBM & XGBoost & FTTransformer \\
elevators & \textbf{0.36} & \textbf{0.08} & 0.05 & 16 599 & 19 & 2986.50 & TabNet & XGBoost & CatBoost \\
guillermo & \textbf{0.35} & \textbf{0.60} & 0.00 & 20 000 & 4 297 & NaN & XGBoost & RandomForest & TabNet \\
higgs & \textbf{0.41} & \textbf{0.10} & 0.07 & 98 050 & 29 & 15.53 & ResNet & XGBoost & LightGBM \\
nomao & \textbf{0.22} & \textbf{0.18} & 0.00 & 34 465 & 119 & 1100.34 & LightGBM & XGBoost & CatBoost \\
100-plants-texture & \textbf{0.20} & \textbf{0.11} & 0.00 & 1 599 & 65 & 17.66 & CatBoost & XGBoost & ResNet \\
poker-hand & \textbf{0.58} & \textbf{0.98} & 0.00 & 1 025 009 & 11 & 0.08 & XGBoost & CatBoost & KNN \\
profb & \textbf{0.39} & \textbf{0.38} & 0.00 & 672 & 10 & 0.95 & CatBoost & DeepFM & MLP-rtdl \\
socmob & \textbf{0.24} & \textbf{0.10} & 0.00 & 1 156 & 6 & NaN & XGBoost & CatBoost & ResNet \\
audiology & \textbf{0.43} & 0.03 & 0.00 & 226 & 70 & NaN & STG & XGBoost & ResNet \\
splice & \textbf{0.30} & 0.03 & 0.00 & 3 190 & 61 & NaN & LightGBM & XGBoost & CatBoost \\
vehicle & 0.05 & \textbf{0.10} & \textbf{0.10} & 846 & 19 & 15.16 & TabPFN & SVM & DANet \\
Australian & 0.15 & \textbf{0.08} & 0.00 & 690 & 15 & 2.00 & CatBoost & XGBoost & TabPFN \\
Bioresponse & 0.07 & \textbf{0.07} & 0.00 & 3 751 & 1 777 & 328.77 & LightGBM & XGBoost & CatBoost \\
GesturePhase & 0.08 & \textbf{0.08} & 0.00 & 9 872 & 33 & 52.18 & LightGBM & XGBoost & CatBoost \\
SpeedDating & 0.18 & \textbf{0.14} & 0.00 & 8 378 & 121 & 36.43 & XGBoost & CatBoost & LightGBM \\
ada-agnostic & 0.12 & \textbf{0.11} & 0.00 & 4 562 & 49 & NaN & XGBoost & CatBoost & LightGBM \\
airlines & \textbf{0.20} & \textbf{0.18} & 0.00 & 539 382 & 8 & 2.01 & LightGBM & XGBoost & CatBoost \\
artificial-characters & 0.13 & \textbf{0.11} & 0.00 & 10 218 & 8 & 0.63 & XGBoost & LightGBM & CatBoost \\
colic & 0.13 & \textbf{0.11} & 0.00 & 368 & 27 & 4.00 & CatBoost & XGBoost & FTTransformer \\
credit-approval & 0.12 & \textbf{0.08} & 0.00 & 690 & 16 & 74.77 & CatBoost & TabPFN & XGBoost \\
heart-h & 0.10 & \textbf{0.07} & 0.08 & 294 & 14 & NaN & DeepFM & TabTransformer & NAM \\
jasmine & 0.13 & \textbf{0.13} & 0.00 & 2 984 & 145 & 47.60 & CatBoost & XGBoost & LightGBM \\
kc1 & 0.14 & \textbf{0.07} & 0.00 & 2 109 & 22 & 28.34 & CatBoost & XGBoost & FTTransformer \\
lymph & 0.14 & \textbf{0.08} & 0.00 & 148 & 19 & 17.04 & XGBoost & DANet & SAINT \\
mfeat-fourier & 0.00 & \textbf{0.07} & 0.07 & 2 000 & 77 & 0.64 & SVM & SAINT & STG \\
phoneme & 0.10 & \textbf{0.15} & 0.00 & 5 404 & 6 & 1.23 & XGBoost & LightGBM & RandomForest \\
qsar-biodeg & 0.08 & \textbf{0.08} & 0.05 & 1 055 & 42 & 93.24 & TabPFN & CatBoost & SAINT \\
balance-scale & 0.07 & 0.05 & \textbf{0.16} & 625 & 5 & 0.02 & TabPFN & SAINT & MLP \\
cnae-9 & 0.11 & 0.04 & \textbf{0.10} & 1 080 & 857 & NaN & TabTransformer & STG & MLP-rtdl \\
mfeat-zernike & 0.00 & 0.04 & \textbf{0.10} & 2 000 & 48 & 1.42 & SVM & DANet & ResNet \\
monks-problems-2 & 0.04 & 0.00 & \textbf{0.17} & 601 & 7 & NaN & SAINT & ResNet & MLP-rtdl \\
\bottomrule
\end{tabular}
}
\end{table}

\paragraph{Hard for most algorithms.} 
This criterion is designed to include datasets on which most algorithms were not able to reach top performance.
In particular, a dataset satisfies the criterion if the fourth-best log loss out of \nalgs{} algorithms is at least 7\% worse than the top log loss.
In other words, this criterion will include datasets on which one, two, or three algorithms were able to stand out in terms of performance.
For example, if ten algorithms were all able to achieve a performance within 7\% of the top-performing algorithm, we can reasonably assume that the dataset might not be `hard'.
Interestingly, this criterion subsumes the majority of the datasets from the previous criterion.

\paragraph{Hard for GBDTs.}
The first two criteria result in datasets such that GBDTs primarily are the top-performing methods. This is not surprising in light of the overall performance of GBDTs that we showed in \cref{sec:analysis}.
However, for the field of tabular data to progress, focusing on datasets for which GBDTs already perform well would leave a blindspot on datasets for which GBDTs perform poorly.
Therefore, we add all datasets for which GBDTs perform 10\% worse than the top-performing algorithm, in order to achieve a greater diversity of datasets.

\paragraph{TabZilla Characteristics.}
\cref{tab:benchmark-datasets} shows all datasets, their statistics, and their top three algorithms.
Based on the criteria alone, the dataset characteristics are diverse, with sizes ranging from 148 to over 1 million, as well as a large range of the variance of kurtosis of the features (one of our measures of irregularity).

In \cref{app:tabzilla}, we compare the performance of all algorithms on the benchmark
291 suite. 
In order to accelerate research in tabular data, we release TabZilla as a collection in OpenML, and we open-source all of our computed metafeatures and results.
In \cref{app:documentation}, we give the full dataset documentation, including a datasheet \citep{gebru2021datasheets}.

%% file: conclusion.tex
\section{Conclusions and Future Work}\label{sec:conclusion}

In this work, we conducted the largest tabular data analysis to date, by comparing \nalgs{} approaches across \ndatasets{} datasets.
We found that the `NN vs.\ GBDT' debate is overemphasized: for a surprisingly high number of datasets, either a simple baseline method performs on par with all other methods, or light hyperparameter tuning on a GBDT increases performance more than choosing the best algorithm.
On the other hand, on average, GBDTs do outperform NNs.
We also analyzed what properties of a dataset make NNs or GBDTs better-suited to perform well. For example, GBDTs are better than NNs at handling various types of data irregularity.
Finally, based on our analysis, we released TabZilla, a collection of the \nsuite{} `hardest' out of the \ndatasets{} datasets we studied: hard for baselines, most algorithms, and  GBDTs.
The goal in releasing TabZilla is to accelerate tabular data research by focusing on improving the current blind spots in the literature.

Our work provides a large set of tools to accelerate research on tabular data.
For example, researchers developing a new neural net for tabular data can use our open-source repository to immediately compare their method to \nalgs{} algorithms across \ndatasets{} datasets.
Researchers can also use our metafeature analysis to improve the weaknesses of current or future algorithms; for example, making neural nets more robust to data irregularities is a natural next step.
Furthermore, researchers studying ensemble methods can weight the models differently for each dataset based on its metafeatures.
Finally, our open-source collection of extensive datasets and metafeatures can make it easier for researchers to design new meta-learned \citep{hollmann2022tabpfn} or pre-trained models for tabular data \citep{levin2022transfer,hegselmann2023tabllm,zhu2023xtab}.
There are several interesting ideas for extensions such as regression datasets, time-series forecasting datasets, studying uncertainty quantification, studying the effect of the percentage of categorical features on NNs, and studying more comprehensive hyperparameter optimization including regularization methods.

%% file: impact.tex
\section{Broader Societal Impact Statement} \label{app:impact}

The goal of our work is to conduct an analysis of tabular data, including the significance of `GBDTs vs.\ NNs', as well as to provide a large set of tools for researchers and practitioners working on tabular data.
We do not see any negative broader societal impacts of our work.
In fact, our work shows that on a surprisingly large fraction of datasets, it is not necessary to train a resource-intensive neural net: a simple baseline, or tuning CatBoost, is enough to reach top performance.
We even predict which datasets are more amenable to GBDTs: larger datasets, datasets with a high size-to-number-of-features ratio, and `irregular' datasets.
Our hope is that our work will have a positive impact for both practitioners and researchers: by providing the largest analysis and open-source codebase to date, along with a benchmark suite of `hard' datasets, our work can both accelerate future research and make the comparisons in future work more rigorous and comprehensive.

%% file: appendix_documentation.tex
\section{Dataset Documentation} \label{app:documentation}

In this section, we give an overview of the documentation for our dataset. 
For the full details, including usage and tutorials, see
\url{https://github.com/naszilla/tabzilla}.

\subsection{Author responsibility and license}
We, the authors, bear all responsibility in case of violation of rights. 
The license of our repository is the \textbf{Apache License 2.0}. For more information, see \url{https://github.com/naszilla/tabzilla/blob/main/LICENSE}.

\subsection{Maintenance plan}

The data is available on OpenML at \url{https://www.openml.org/search?type=study&study_type=task&id=379&sort=tasks_included}.

We plan to actively maintain the benchmark suite, and we welcome contributions from the community. 

\subsection{Code of conduct}
Our Code of Conduct is from the Contributor Covenant, version 2.0. See \\
\url{https://www.contributor-covenant.org/version/2/0/code_of_conduct.html}. \\

\subsection{Datasheet}

We include a datasheet \citep{gebru2021datasheets} for TabZilla, which is available here and also at \url{https://github.com/naszilla/tabzilla}.

\input{datasheet}

%% file: datasheet.tex

\subsubsection*{Motivation For Datasheet Creation}

\textcolor{blue}{\subsubsubsection*{Why was the datasheet created? (e.g., was there a specific task in mind? was there a specific gap that needed to
be filled?)}}
The goal of releasing the TabZilla Benchmark Suite is to accelerate research in tabular data by introducing a set of `hard' datasets.
Specifically, simple baselines cannot reach top performance, and most algorithms (out of the \nalgs{} we tried) cannot reach top performance.
We found that a surprisingly high percentage of datasets used in tabular research today are such that a simple baseline can reach just as high accuracy as the leading methods.

\textcolor{blue}{\subsubsubsection*{Has the dataset been used already? If so, where are the results so others can compare
(e.g., links to published papers)?}}
All of the individual datasets are already released in OpenML, and many have been used in prior work on tabular data. 
However, our work gathers these datasets into a single `hard' suite.

\textcolor{blue}{\subsubsubsection*{What (other) tasks could the dataset be used for?}}
All of these datasets are tabular classification datasets, and so to the best of our knowledge, they cannot be used for anything other than tabular classification.

\textcolor{blue}{\subsubsubsection*{Who funded the creation dataset?}}
This benchmark suite was created by researchers at Abacus.AI, Stanford, Pinterest, University of Maryland, IIT Bombay, New York University, and Caltech. Funding for the dataset computation itself is from Abacus.AI.

\textcolor{blue}{\subsubsubsection*{Any other comment?}}
None.

\subsubsection*{Datasheet Composition}

\textcolor{blue}{\subsubsubsection*{What are the instances?(that is, examples; e.g., documents, images, people, countries) Are there multiple types
of instances? (e.g., movies, users, ratings; people, interactions between them; nodes, edges)}}
Each instance is a tabular datapoint.
The makeup of each point depends on its dataset.
For example, three of the datasets consist of poker hands, electricity usage, and plant textures.

\textcolor{blue}{\subsubsubsection*{How many instances are there in total (of each type, if appropriate)?}}
See Table \ref{tab:benchmark-datasets} for a breakdown of the number of instances for each dataset.

\textcolor{blue}{\subsubsubsection*{What data does each instance consist of ? “Raw”
data (e.g., unprocessed text or images)? Features/attributes? Is there a label/target associated with
instances? If the instances related to people, are subpopulations identified (e.g., by age, gender, etc.) and what is
their distribution?}}
The raw data is hosted on OpenML. In our repository, we also contain scripts for the standard preprocessing we ran before training tabular data models.
The data are not related to people.


\textcolor{blue}{\subsubsubsection*{Is any information missing from individual instances? If so, please provide a description, explaining why this information is missing (e.g., because it was unavailable). This does not include intentionally removed information, but might include, e.g., redacted text.}}
There is no missing information.

\textcolor{blue}{\subsubsubsection*{Are relationships between individual instances made explicit (e.g.,
users’ movie ratings, social network links)? If so, please describe
how these relationships are made explicit.}}
There are no relationships between individual instances.

\textcolor{blue}{\subsubsubsection*{Does the dataset contain all possible instances or is it a sample (not
necessarily random) of instances from a larger set? If the dataset is
a sample, then what is the larger set? Is the sample representative of the
larger set (e.g., geographic coverage)? If so, please describe how this
representativeness was validated/verified. If it is not representative of the
larger set, please describe why not (e.g., to cover a more diverse range of
instances, because instances were withheld or unavailable).}}

We selected the datasets for our benchmark suite as follows.
We started with \ndatasets{} datasets, which we selected with the aim to include most classification datasets from popular recent papers that study tabular data \citep{kadra2021well,borisov2021deep,shwartz2022tabular,gorishniy2021revisiting}, including datasets from the OpenML-CC18 suite \cite{bischl2017openml}, the OpenML Benchmarking Suite \cite{gijsbers2019open}, and additional OpenML datasets \citep{vanschoren2014openml}.
Due to the scale of our experiments (\nmodels{} total models trained), we limited to datasets smaller than 1.1M.
CC-18 and OpenML Benchmarking Suite are both seen as the go-to standards for conducting a fair, diverse evaluation across algorithms due to their rigorous selection criteria and wide diversity of datasets \citep{bischl2017openml,gijsbers2019open}.
Out of these \ndatasets{} datasets, we selected \nsuite{} datasets for our suite as described in \cref{sec:tabzilla}.

\textcolor{blue}{\subsubsubsection*{Are there recommended data splits (e.g., training, development/validation, testing)? If so, please provide a description of these
splits, explaining the rationale behind them.}}

We use the 10 folds from OpenML, and it is recommended to report performance averaged over these 10 folds, as we do and as OpenML does.
If a validation set is required, we recommend additionally using the validation splits that we used, described in \cref{sec:analysis}.

\textcolor{blue}{\subsubsubsection*{Are there any errors, sources of noise, or redundancies in the
dataset? If so, please provide a description.}}
There are no known errors, sources of noise, or redundancies.

\textcolor{blue}{\subsubsubsection*{Is the dataset self-contained, or does it link to or otherwise rely on
external resources (e.g., websites, tweets, other datasets)? If it links
to or relies on external resources, a) are there guarantees that they will
exist, and remain constant, over time; b) are there official archival versions
of the complete dataset (i.e., including the external resources as they existed at the time the dataset was created); c) are there any restrictions
(e.g., licenses, fees) associated with any of the external resources that
might apply to a future user? Please provide descriptions of all external
resources and any restrictions associated with them, as well as links or
other access points, as appropriate.}}
The dataset is self-contained.

\textcolor{blue}{Any other comments?}
None.

\subsubsection*{Collection Process}

\textcolor{blue}{\subsubsubsection*{What mechanisms or procedures were used to collect the data (e.g.,
hardware apparatus or sensor, manual human curation, software program, software API)? How were these mechanisms or procedures validated?}}
We did not create the individual datasets.
However, we selected the datasets for our benchmark suite as follows.
We started with \ndatasets{} datasets, which we selected with the aim to include most classification datasets from popular recent papers that study tabular data \citep{kadra2021well,borisov2021deep,shwartz2022tabular,gorishniy2021revisiting}, including datasets from the OpenML-CC18 suite \cite{bischl2017openml}, the OpenML Benchmarking Suite \cite{gijsbers2019open}, and additional OpenML datasets \citep{vanschoren2014openml}.
Due to the scale of our experiments (\nmodels{} total models trained), we limited to datasets smaller than 1.1M.
CC-18 and OpenML Benchmarking Suite are both seen as the go-to standards for conducting a fair, diverse evaluation across algorithms due to their rigorous selection criteria and wide diversity of datasets \citep{bischl2017openml,gijsbers2019open}.

Out of these \ndatasets{} datasets, we selected \nsuite{} datasets for our suite as described in \cref{sec:tabzilla}.

\textcolor{blue}{\subsubsubsection*{How was the data associated with each instance acquired? Was the
data directly observable (e.g., raw text, movie ratings), reported by subjects (e.g., survey responses), or indirectly inferred/derived from other data
(e.g., part-of-speech tags, model-based guesses for age or language)?
If data was reported by subjects or indirectly inferred/derived from other
data, was the data validated/verified? If so, please describe how.}}
The datasets were selected using the three criteria from \cref{sec:tabzilla}.

\textcolor{blue}{\subsubsubsection*{If the dataset is a sample from a larger set, what was the sampling strategy (e.g., deterministic, probabilistic with specific sampling probabilities)?}}
As described earlier, the datasets were selected using the three criteria from \cref{sec:tabzilla}.

\textcolor{blue}{\subsubsubsection*{Who was involved in the data collection process (e.g., students,
crowdworkers, contractors) and how were they compensated (e.g.,
how much were crowdworkers paid)?}}
The creation of the TabZilla Benchmark Suite was done by the authors of this work.

\textcolor{blue}{\subsubsubsection*{Over what timeframe was the data collected? Does this timeframe
match the creation timeframe of the data associated with the instances
(e.g., recent crawl of old news articles)? If not, please describe the timeframe in which the data associated with the instances was created.}}
The timeframe for constructing the TabZilla Benchmark Suite was from April 15, 2023 to June 1, 2023.

\subsubsection*{Data Preprocessing}

\textcolor{blue}{\subsubsubsection*{Was any preprocessing/cleaning/labeling of the data done (e.g., discretization or bucketing, tokenization, part-of-speech tagging, SIFT
feature extraction, removal of instances, processing of missing values)? If so, please provide a description. If not, you may skip the remainder of the questions in this section.}}

We include both the raw data and the preprocessed data.
We preprocessed the data by imputing each NaN to the mean of the respective feature.
We left all other preprocessing (such as scaling) to the algorithms themselves.

\textcolor{blue}{\subsubsubsection*{Was the “raw” data saved in addition to the preprocessed/cleaned/labeled data (e.g., to support unanticipated future uses)? If so, please provide a link or other access point to the “raw” data.}}

The raw data is available at 
\url{https://www.openml.org/search?type=study&study_type=task&id=379&sort=tasks_included}.

\textcolor{blue}{\subsubsubsection*{Is the software used to preprocess/clean/label the instances available? If so, please provide a link or other access point.}}

Our README contains an extensive section on the data preprocessing, here:
\url{https://github.com/naszilla/tabzilla#openml-datasets}.

\textcolor{blue}{\subsubsubsection*{Does this dataset collection/processing procedure achieve the motivation for creating the dataset stated in the first section of this datasheet? If not,
what are the limitations?}}
We hope that the release of this benchmark suite will achieve our goal of accelerating research in tabular data, as well as making it easier for researchers and practitioners to devise and compare algorithms.
Time will tell whether our suite will be adopted by the community.

\textcolor{blue}{\subsubsubsection*{Any other comments}}
None.

\subsubsection*{Dataset Distribution}

\textcolor{blue}{\subsubsubsection*{How will the dataset be distributed? (e.g., tarball on
website, API, GitHub; does the data have a DOI and is it
archived redundantly?)}}
The benchmark suite is on OpenML at \url{https://www.openml.org/search?type=study&study_type=task&id=379&sort=tasks_included}.

\textcolor{blue}{\subsubsubsection*{When will the dataset be released/first distributed?
What license (if any) is it distributed under?}}
The benchmark suite is public as of June 1, 2023, distributed under the Apache License 2.0.

\textcolor{blue}{\subsubsubsection*{Are there any copyrights on the data?}}
There are no copyrights on the data.

\textcolor{blue}{\subsubsubsection*{Are there any fees or access/export restrictions?}}
There are no fees or restrictions.

\textcolor{blue}{\subsubsubsection*{Any other comments?}}
None.

\subsubsection*{Dataset Maintenance}

\textcolor{blue}{\subsubsubsection*{Who is supporting/hosting/maintaining the
dataset?}}
The authors of this work are supporting/hosting/maintaining the dataset.

\textcolor{blue}{\subsubsubsection*{Will the dataset be updated? If so, how often and
by whom?}}
We welcome updates from the tabular data community. If new algorithms are created, the authors may open a pull request to include their method.

\textcolor{blue}{\subsubsubsection*{How will updates be communicated? (e.g., mailing
list, GitHub)}}
Updates will be communicated on the GitHub README \url{https://github.com/naszilla/tabzilla}.

\textcolor{blue}{\subsubsubsection*{If the dataset becomes obsolete how will this be
communicated?}}
If the dataset becomes obsolete, it will be communicated on the GitHub README \url{https://github.com/naszilla/tabzilla}.

%

\textcolor{blue}{\subsubsubsection*{If others want to extend/augment/build on this
dataset, is there a mechanism for them to do so?
If so, is there a process for tracking/assessing the
quality of those contributions. What is the process
for communicating/distributing these contributions
to users?}}
Others can create a pull request on GitHub with possible extensions to our benchmark suite, which will be approved case-by-case. 
For example, an author of a new hard tabular dataset may create a PR in our codebase with the new dataset. 
These updates will again be communicated on the GitHub README.

\subsubsection*{Legal and Ethical Considerations}

\textcolor{blue}{\subsubsubsection*{Were any ethical review processes conducted (e.g., by an institutional review board)? If so, please provide a description of these review
processes, including the outcomes, as well as a link or other access point
to any supporting documentation.}}
There was no ethical review process. We note that our benchmark suite consists of existing datasets that are already publicly available on OpenML.

\textcolor{blue}{\subsubsubsection*{Does the dataset contain data that might be considered confidential
(e.g., data that is protected by legal privilege or by doctorpatient confidentiality, data that includes the content of individuals non-public
communications)? If so, please provide a description.}}
The datasets do not contain any confidential data.

\textcolor{blue}{\subsubsubsection*{Does the dataset contain data that, if viewed directly, might be offensive, insulting, threatening, or might otherwise cause anxiety? If so,
please describe why}}
None of the data might be offensive, insulting, threatening, or otherwise cause anxiety.

\textcolor{blue}{\subsubsubsection*{Does the dataset relate to people? If not, you may skip the remaining
questions in this section.}}
The datasets do not relate to people.

\textcolor{blue}{\subsubsubsection*{Any other comments?}}
None.

%% file: appendix_related_work.tex
\section{Additional Related Work} \label{app:relatedwork}

\paragraph{Gradient-boosted decision trees.}
GBDTs iteratively build an ensemble of decision trees, with each new tree fitting the residual of the loss from the previous trees, using gradient descent to minimize the losses.
GBDTs have been a powerful tool for modeling tabular data ever since their creation in 2001 \citep{friedman2001greedy}, and numerous works propose high-performing GBDT variants.
XGBoost (eXtreme Gradient Boosting) \citep{chen2016xgboost} uses weighted quantile sketching and sparsity-awareness, allowing it to scale to large datasets.
LightGBM (Light Gradient Boosting Machine) \citep{ke2017lightgbm} uses gradient-based one-sided sampling and exclusive feature bundling to create a faster and more lightweight GBDT implementation.
CatBoost (Categorical Boosting) \citep{prokhorenkova2018catboost} introduces ordered boosting, a new method for handling categorical features, as well as better methods for handling missing values and outliers.

\paragraph{Neural networks for tabular data.}
In their survey on deep learning for tabular data, 
Borisov et al.\ described three types of tabular data approaches for neural networks \citep{borisov2021deep}.
Data transformation methods \citep{yoon2020vime,hancock2020survey} seek to encode the data into a format that is better-suited for neural nets.
Architecture-based methods design specialized neural architectures for tabular data \citep{Popov2020Neural,guo2017deepfm,chen2022danets}, a large sub-class of which are transformer-based architectures \citep{arik2021tabnet,gorishniy2021revisiting,huang2020tabtransformer,somepalli2021saint}.
Regularization-based methods specially tailor regularizers methods to improve the performance of a given architecture \citep{shavitt2018regularization,kadra2021well,jeffares2022tangos}.
Notably, one recent work designs a new framework for regularization in the tabular
setting built on latent unit attributions \citep{jeffares2022tangos}, and another recent work shows that searching for the optimal combination/cocktail of 13 regularization techniques applied to a simple neural net achieves strong performance.
While regularization was not the focus of our current work, including regularization methods would be an exciting direction for follow-up work.

A notable recent neural network for tabular data is TabPFN \citep{hollmann2022tabpfn,hollmann2023gpt}, a prior-data fitted network \citep{muller2021transformers,nagler2023statistical} for tabular data.
TabPFN is a \emph{meta-learned} algorithm that provably approximates Bayesian inference, which can make predictions on a new dataset in under one second.
PFNs have recently been applied to learning curve extrapolation \citep{adriaensen2022efficient}, forecasting \citep{khurana2023forecastpfn}, and Bayesian optimization \citep{muller2023pfns4bo}.

\paragraph{GBDTs versus NNs.}
Several recent works compare GBDTs to NNs on tabular data, often finding that \emph{either} neural nets \citep{kadra2021well,gorishniy2021revisiting,arik2021tabnet,Popov2020Neural}
or GBDTs \citep{shwartz2022tabular,borisov2021deep,grinsztajn2022tree,gorishniy2021revisiting}
perform best.
Shwartz-Ziv and Armon compare GBDTs and NNs on 30 datasets, finding that GBDTs perform better on average, and ensembling both achieves better performance \citep{shwartz2022tabular}. 
Kadra et al.\ compare GBDTs and NNs on 40 datasets, finding that properly-tuned neural networks perform best on average \citep{kadra2021well}.

Gorishniy et al.\ \citep{gorishniy2021revisiting} introduce a ResNet-like \citep{resnet} architecture, and FT-Transformer, a transformer-based \citep{vaswani2017attention} architecture.
Across experiments on eleven datasets, they conclude that there is still no universal winner among GBDTs and NNs.
Borisov et al.\ \citep{borisov2021deep} compare classical machine learning methods with eleven deep learning approaches on five tabular datasets, concluding that GBDTs still have the edge.
In the transfer learning setting, Levin et al.\ \citep{levin2023transfer} find that neural networks have a decisive edge over GBDTs when pre-training data is available.

Perhaps the  most related work to ours is by Grinsztajn et al.\ \citep{grinsztajn2022tree}, who investigate why tree-based methods outperform neural nets on tabular data.
There are a few differences between their work and ours.
First, they only consider seven algorithms and 45 datasets, compared to our 19 algorithms and 176 datasets.
Second, their dataset sizes range from 3\,000 to 10\,000, or seven that are exactly 50\,000, in contrast to our dataset sizes which range from 32 to 1\,025\,009 (see Table \ref{tab:dataset-stats}).
Additionally, they further control their study, for example by upper bounding the ratio of size to features, by removing high-cardinality categorical feature, and by removing low-cardinality numerical features. While this has the benefit of being a more controlled study, their analysis misses out on some of our observations, such as GBDTs performing better than NNs on `irregular' datasets.
Finally, while Grinsztajn et al.\ focused in depth on a few metafeatures such as dataset smoothness and number of uninformative features, our work considers orders of magnitude more metafeatures. Again, while each approach has its own strengths, our work is able to discover more potential insights, correlations, and takeaways for practitioners.

%% file: appendix_analysis.tex
\section{Additional Experiments} \label{app:experiments}

We give additional experiments, including dataset statistics (\cref{app:dataset-stats}), additional results from \cref{subsec:relative-performance} (\cref{app:relative-performance}), and additional results from \cref{subsec:metafeature} (\cref{app:metafeature}). 

\subsection{Dataset statistics} \label{app:dataset-stats}

\cref{tab:dataset-stats} shows summary statistics for all 176 datasets used in our experiments.
Roughly half of our datasets have a binary classification target (and as many as 100 target classes), and roughly half of all training sets have fewer than 2300 instances---though many datasets have tens of thousands of instances.

\begin{table}[t]
    \centering
    \caption{Summary statistics for all 176 datasets used in our experiments. Left columns show the number of instances, number of features, number of target classes, and the ratio of the mimum frerquency of any class to the maximum frequency of any class (Min-Max Class Freq.). Right columns show the number of feature types. All statistics except for class frequency ratio are rounded to the nearest integer.\label{tab:dataset-stats}}
\begin{tabular}{lrrrrrrrrr}
\toprule
{} &  \# Inst. &  \# Feats. &  \# Classes & Min-Max Class Freq. & \multicolumn{3}{c}{\# Feature Types}  \\\cline{6-8}
{} & & & & &  Num. &  Bin. & Cat.\\ 
\midrule
mean & 30567 & 223 & 6 & 0.48 & 206 & 25 & 17 \\
std & 106943 & 786 & 12 & 0.35 & 781 & 144 & 119 \\
min & 32 & 2 & 2 & 2e-05 & 0 & 0 & 0 \\
25\% & 596 & 9 & 2 & 0.14 & 4 & 0 & 0 \\
50\% & 2218 & 21 & 2 & 0.46 & 10 & 0 & 0 \\
75\% & 11008 & 61 & 6 & 0.82 & 50 & 2 & 8 \\
max & 1025009 & 7200 & 100 & 1.00 & 7200 & 1555 & 1555 \\
\bottomrule
\end{tabular}
\end{table}

\subsection{Additional experiments from \cref{subsec:relative-performance}} \label{app:relative-performance}

In this section, we give additional experiments from \cref{subsec:relative-performance}, including relative performance tables, training time analysis, and critical difference diagrams.

\subsubsection{Relative performance tables} \label{app:relative-perf}

First, we show the ranking of all tuned algorithms according to each performance metric, averaged over datasets: log loss (\cref{tab:updated-ranks-ll}), F1 score (\cref{tab:updated-ranks-f1}), and ROC-AUC (\cref{tab:updated-ranks-auc}).
These are similar to \cref{tab:main-performance}, but with different metrics. 
Rankings are calculated by first averaging tuned performance over all 10 splits of each dataset, and then ranking each algorithm for each dataset according to their average performance. 
For these rankings, a tie between two algorithms is indicated by the \emph{lowest} (best) ranking. So if multiple algorithms achieve the highest average accuracy for a dataset, they both receive ranking 1.

\begin{table}
    \centering
\caption{Performance of algorithms according to Log Loss over 98 datasets. 
Columns show the rank over all datasets, the average normalized log loss (Mean LL), the standard deviation of normalized LL across folds (Std.\ LL), and the train time per 1000 instances. Min/max/mean/median are taken over all datasets.}
\label{tab:updated-ranks-ll}
\resizebox{\linewidth}{!}{
\begin{tabular}{llrrrrrrrrrr}
\toprule
{} & {} & \multicolumn{4}{l}{Rank} & \multicolumn{2}{l}{Mean LL} & \multicolumn{2}{l}{Std.\ LL} & \multicolumn{2}{l}{Time /1000 inst.} \\
Algorithm &   Class &             min & max &   mean & med.\ &                           mean & med.\ &                          mean & med.\ &                              mean &  med.\ \\
\midrule
TabPFN$^*$ & PFN & 1 & 15 & 4.49 & 3 & 0.05 & 0.01 & 0.07 & 0.04 & 0.25 & 0.01 \\
CatBoost & GBDT & 1 & 14 & 4.92 & 4 & 0.04 & 0.02 & 0.08 & 0.06 & 13.89 & 1.66 \\
XGBoost & GBDT & 1 & 15 & 5.31 & 5 & 0.05 & 0.02 & 0.08 & 0.06 & 0.73 & 0.37 \\
SAINT & NN & 1 & 19 & 7.20 & 6 & 0.12 & 0.06 & 0.09 & 0.08 & 202.59 & 173.23 \\
ResNet & NN & 1 & 16 & 8.30 & 8 & 0.12 & 0.08 & 0.10 & 0.08 & 16.12 & 8.97 \\
SVM & base & 1 & 18 & 8.34 & 9 & 0.14 & 0.06 & 0.08 & 0.05 & 49.83 & 1.20 \\
LightGBM & GBDT & 1 & 19 & 8.38 & 7.50 & 0.13 & 0.07 & 0.21 & 0.09 & 0.83 & 0.27 \\
DANet & NN & 1 & 18 & 8.62 & 9 & 0.12 & 0.09 & 0.13 & 0.08 & 71.58 & 61.35 \\
FTTransformer & NN & 1 & 17 & 9.34 & 10 & 0.14 & 0.09 & 0.11 & 0.09 & 29.58 & 18.48 \\
STG & NN & 1 & 18 & 9.68 & 10 & 0.18 & 0.07 & 0.07 & 0.05 & 18.82 & 15.85 \\
RandomForest & base & 1 & 19 & 10.57 & 11 & 0.19 & 0.13 & 0.22 & 0.07 & 0.29 & 0.22 \\
LinearModel & base & 1 & 19 & 10.58 & 11 & 0.24 & 0.10 & 0.10 & 0.06 & 0.04 & 0.03 \\
MLP-rtdl & NN & 1 & 19 & 11.71 & 12 & 0.28 & 0.18 & 0.17 & 0.12 & 13.75 & 7.96 \\
NODE & NN & 1 & 18 & 11.89 & 12 & 0.23 & 0.18 & 0.04 & 0.03 & 196.82 & 176.16 \\
MLP & NN & 1 & 19 & 12.51 & 13 & 0.29 & 0.21 & 0.15 & 0.11 & 18.29 & 10.95 \\
TabNet & NN & 1 & 19 & 12.90 & 14.50 & 0.40 & 0.25 & 0.38 & 0.19 & 34.62 & 29.69 \\
VIME & NN & 3 & 19 & 14.36 & 16 & 0.40 & 0.37 & 0.09 & 0.07 & 16.92 & 14.64 \\
KNN & base & 1 & 19 & 15.32 & 16 & 0.51 & 0.42 & 0.37 & 0.20 & 0.01 & 0.00 \\
DecisionTree & base & 1 & 19 & 15.59 & 18 & 0.56 & 0.58 & 0.54 & 0.41 & 0.02 & 0.01 \\
\bottomrule
\end{tabular}
}
\end{table}

\begin{table}
    \centering
\caption{Performance of algorithms according to F1 score over 98 datasets. 
Columns show the rank over all datasets, the average normalized F1 score (Mean F1), the standard deviation of normalized F1 score across folds (Std.\ F1), and the train time per 1000 instances. Min/max/mean/median are taken over all datasets.}
\label{tab:updated-ranks-f1}
\resizebox{\linewidth}{!}{
\begin{tabular}{llrrrrrrrrrr}
\toprule
{} & {} & \multicolumn{4}{l}{Rank} & \multicolumn{2}{l}{Mean F1} & \multicolumn{2}{l}{Std.\ F1} & \multicolumn{2}{l}{Time /1000 inst.} \\
Algorithm & Class  &            min & max &   mean & med.\ &                           mean & med.\ &                          mean & med.\ &                              mean &  med.\ \\
\midrule
CatBoost & GBDT & 1 & 18 & 5.77 & 4 & 0.87 & 0.93 & 0.29 & 0.22 & 21.00 & 2.08 \\
TabPFN$^*$ & PFN & 1 & 19 & 5.97 & 5 & 0.83 & 0.93 & 0.26 & 0.19 & 0.25 & 0.01 \\
XGBoost & GBDT & 1 & 17 & 6.78 & 6 & 0.82 & 0.89 & 0.32 & 0.22 & 0.83 & 0.37 \\
ResNet & NN & 1 & 19 & 7.67 & 7.50 & 0.76 & 0.82 & 0.29 & 0.19 & 16.04 & 9.34 \\
NODE & NN & 1 & 19 & 8.05 & 8 & 0.75 & 0.81 & 0.25 & 0.19 & 140.71 & 117.04 \\
SAINT & NN & 1 & 19 & 8.07 & 7 & 0.73 & 0.86 & 0.30 & 0.23 & 171.14 & 144.37 \\
FTTransformer & NN & 1 & 17 & 8.17 & 8 & 0.76 & 0.82 & 0.30 & 0.19 & 27.94 & 18.40 \\
RandomForest & base & 1 & 19 & 8.20 & 7 & 0.77 & 0.83 & 0.31 & 0.21 & 0.36 & 0.25 \\
LightGBM & GBDT & 1 & 19 & 8.47 & 8 & 0.76 & 0.83 & 0.35 & 0.21 & 0.86 & 0.31 \\
SVM & base & 1 & 18 & 9.05 & 9.50 & 0.69 & 0.77 & 0.25 & 0.17 & 29.99 & 1.73 \\
DANet & NN & 1 & 18 & 9.69 & 10 & 0.74 & 0.81 & 0.31 & 0.22 & 69.54 & 60.20 \\
MLP-rtdl & NN & 1 & 19 & 9.72 & 10 & 0.66 & 0.73 & 0.27 & 0.16 & 14.29 & 7.30 \\
DecisionTree & base & 1 & 19 & 11.69 & 13 & 0.61 & 0.71 & 0.34 & 0.24 & 0.03 & 0.01 \\
STG & NN & 1 & 19 & 11.81 & 12 & 0.57 & 0.65 & 0.28 & 0.18 & 18.43 & 15.76 \\
MLP & NN & 1 & 19 & 12.18 & 13 & 0.57 & 0.58 & 0.29 & 0.18 & 18.42 & 11.20 \\
LinearModel & base & 1 & 19 & 12.24 & 14 & 0.52 & 0.53 & 0.30 & 0.24 & 0.04 & 0.03 \\
TabNet & NN & 1 & 19 & 12.66 & 14 & 0.56 & 0.61 & 0.38 & 0.26 & 34.82 & 29.16 \\
KNN & base & 1 & 19 & 13.60 & 15 & 0.46 & 0.55 & 0.28 & 0.21 & 0.01 & 0.00 \\
VIME & NN & 2 & 19 & 15.11 & 17 & 0.36 & 0.34 & 0.27 & 0.18 & 17.02 & 14.96 \\
\bottomrule
\end{tabular}
}
\end{table}

\begin{table}
    \centering
\caption{Performance of algorithms according to ROC-AUC, over 98 datasets. 
Columns show the rank over all datasets, the average normalized ROC-AUC score (Mean AUC), the standard deviation of normalized ROC-AUC score across folds (Std.\ AUC), and the train time per 1000 instances. Min/max/mean/median are taken over all datasets.}
\label{tab:updated-ranks-auc}
\resizebox{\linewidth}{!}{
\begin{tabular}{llrrrrrrrrrr}
\toprule
{} & {} & \multicolumn{4}{l}{Rank} & \multicolumn{2}{l}{Mean AUC} & \multicolumn{2}{l}{Std.\ AUC} & \multicolumn{2}{l}{Time /1000 inst.} \\
Algorithm &  Class &              min & max &   mean & med.\ &                           mean & med.\ &                          mean & med.\ &                              mean &  med.\ \\
\midrule
TabPFN$^*$ & PFN & 1 & 16 & 4.87 & 3 & 0.91 & 0.97 & 0.17 & 0.07 & 0.25 & 0.01 \\
CatBoost & GBDT & 1 & 18 & 5.64 & 4 & 0.92 & 0.96 & 0.18 & 0.08 & 20.51 & 1.94 \\
XGBoost & GBDT & 1 & 17 & 6.12 & 5 & 0.90 & 0.96 & 0.19 & 0.09 & 0.84 & 0.38 \\
ResNet & NN & 1 & 19 & 7.11 & 6 & 0.84 & 0.93 & 0.19 & 0.12 & 15.83 & 8.78 \\
SAINT & NN & 1 & 19 & 7.22 & 6 & 0.84 & 0.93 & 0.19 & 0.12 & 170.31 & 145.99 \\
RandomForest & base & 1 & 18 & 7.45 & 7 & 0.87 & 0.94 & 0.19 & 0.10 & 0.41 & 0.28 \\
DANet & NN & 1 & 17 & 8.08 & 8 & 0.83 & 0.91 & 0.19 & 0.08 & 64.15 & 57.12 \\
LightGBM & GBDT & 1 & 19 & 8.69 & 8 & 0.83 & 0.91 & 0.23 & 0.10 & 0.89 & 0.29 \\
NODE & NN & 1 & 19 & 9.17 & 10 & 0.81 & 0.90 & 0.20 & 0.14 & 160.58 & 131.56 \\
FTTransformer & NN & 1 & 19 & 9.40 & 9.50 & 0.79 & 0.88 & 0.20 & 0.13 & 27.73 & 18.00 \\
SVM & base & 1 & 19 & 9.87 & 10.50 & 0.75 & 0.87 & 0.22 & 0.10 & 61.16 & 2.01 \\
MLP-rtdl & NN & 1 & 19 & 10.02 & 10 & 0.73 & 0.83 & 0.21 & 0.12 & 15.05 & 7.01 \\
STG & NN & 1 & 19 & 11.33 & 12 & 0.66 & 0.79 & 0.24 & 0.15 & 18.58 & 15.98 \\
LinearModel & base & 1 & 19 & 12.01 & 13 & 0.62 & 0.75 & 0.23 & 0.17 & 0.04 & 0.03 \\
MLP & NN & 1 & 19 & 12.55 & 13.50 & 0.65 & 0.71 & 0.23 & 0.14 & 18.17 & 11.16 \\
TabNet & NN & 1 & 19 & 12.81 & 14 & 0.63 & 0.75 & 0.32 & 0.19 & 35.06 & 29.32 \\
DecisionTree & base & 1 & 19 & 13.88 & 15 & 0.53 & 0.59 & 0.30 & 0.23 & 0.02 & 0.01 \\
KNN & base & 1 & 19 & 14.59 & 16 & 0.52 & 0.56 & 0.25 & 0.19 & 0.01 & 0.00 \\
VIME & NN & 1 & 19 & 14.98 & 16 & 0.49 & 0.50 & 0.30 & 0.20 & 17.84 & 15.55 \\
\bottomrule
\end{tabular}
}
\end{table}

Some algorithms perform well even \emph{without} hyperparameter tuning. 
Next, we calculate the ranking of all datasets using the same procedure as in \cref{tab:main-performance} and the previous tables, but with their default hyperparameter set displayed as a separate algorithm.
We show ranking results for test accuracy.
See \cref{tab:updated-ranks-accuracy-with-default}.
Many of the best-performing algorithms, including CatBoost, XGBoost, LightGBM, and ResNet, perform fairly well both with and without hyperparameter tuning.

\begin{table}
    \centering
{\small
\caption{Performance of algorithms including algorithms parameterized with their default hyperparameters, according to accuracy, over 104 datasets. 
Columns show the rank over all datasets, the average normalized accuracy (Mean Acc.), the standard deviation of normalized accuracy across folds (Std.\ Acc.), and the train time per 1000 instances. Min/max/mean/median are taken over all datasets.}
\label{tab:updated-ranks-accuracy-with-default}
\begin{tabular}{lrrrrrrrrrr}
\toprule
{} & \multicolumn{4}{l}{Rank} & \multicolumn{2}{l}{Mean Acc.} & \multicolumn{2}{l}{Std.\ Acc.} & \multicolumn{2}{l}{Time /1000 inst.} \\
Algorithm &                min & max &   mean & med.\ &                           mean & med.\ &                          mean & med.\ &                              mean &  med.\ \\
\midrule
CatBoost                &                  1 &  30 &   8.05 &    6.0 &                           0.91 &   0.95 &                          0.20 &   0.12 &                             30.27 &    2.22 \\
CatBoost (default)      &                  1 &  31 &  10.12 &    7.0 &                           0.87 &   0.93 &                          0.20 &   0.12 &                             21.95 &    1.53 \\
XGBoost                 &                  1 &  30 &  10.77 &    8.5 &                           0.87 &   0.93 &                          0.22 &   0.13 &                              0.94 &    0.42 \\
ResNet                  &                  1 &  34 &  11.51 &   11.5 &                           0.84 &   0.90 &                          0.20 &   0.11 &                             16.07 &   10.04 \\
XGBoost (default)       &                  1 &  34 &  12.02 &   10.5 &                           0.85 &   0.91 &                          0.22 &   0.13 &                              1.20 &    0.61 \\
NODE                    &                  1 &  33 &  12.17 &   12.0 &                           0.82 &   0.88 &                          0.18 &   0.12 &                            146.89 &  118.94 \\
SAINT                   &                  1 &  34 &  12.27 &   10.0 &                           0.81 &   0.91 &                          0.20 &   0.15 &                            168.14 &  144.84 \\
FTTransformer           &                  1 &  31 &  12.78 &   13.0 &                           0.83 &   0.87 &                          0.21 &   0.14 &                             29.47 &   18.63 \\
RandomForest            &                  1 &  33 &  13.00 &   12.0 &                           0.83 &   0.88 &                          0.21 &   0.15 &                              0.36 &    0.25 \\
LightGBM                &                  1 &  34 &  13.19 &   12.0 &                           0.85 &   0.90 &                          0.23 &   0.14 &                              1.06 &    0.37 \\
ResNet (default)        &                  1 &  35 &  13.94 &   14.0 &                           0.79 &   0.87 &                          0.22 &   0.14 &                             15.26 &    8.67 \\
SVM                     &                  1 &  32 &  14.24 &   16.0 &                           0.78 &   0.88 &                          0.18 &   0.13 &                             29.22 &    1.37 \\
LightGBM (default)      &                  1 &  34 &  14.38 &   12.5 &                           0.80 &   0.88 &                          0.23 &   0.16 &                              1.30 &    0.50 \\
SAINT (default)         &                  1 &  35 &  14.48 &   12.0 &                           0.76 &   0.88 &                          0.20 &   0.15 &                            135.22 &  107.82 \\
NODE (default)          &                  1 &  34 &  14.61 &   15.0 &                           0.77 &   0.86 &                          0.18 &   0.12 &                             67.54 &   50.85 \\
DANet                   &                  1 &  33 &  15.08 &   15.0 &                           0.83 &   0.87 &                          0.21 &   0.15 &                             69.29 &   60.15 \\
RandomForest (default)  &                  1 &  35 &  16.22 &   15.0 &                           0.76 &   0.82 &                          0.20 &   0.14 &                              0.50 &    0.39 \\
MLP-rtdl                &                  1 &  34 &  16.28 &   18.0 &                           0.74 &   0.83 &                          0.18 &   0.11 &                             14.33 &    7.62 \\
FTTransformer (default) &                  1 &  35 &  18.01 &   20.0 &                           0.71 &   0.80 &                          0.21 &   0.14 &                             25.92 &   15.35 \\
STG                     &                  1 &  34 &  18.73 &   19.0 &                           0.70 &   0.81 &                          0.20 &   0.11 &                             18.47 &   16.00 \\
DecisionTree            &                  1 &  35 &  18.93 &   20.0 &                           0.73 &   0.79 &                          0.23 &   0.18 &                              0.03 &    0.01 \\
SVM (default)           &                  1 &  35 &  19.63 &   24.0 &                           0.64 &   0.74 &                          0.19 &   0.12 &                              1.09 &    0.38 \\
MLP                     &                  1 &  34 &  19.66 &   21.5 &                           0.69 &   0.73 &                          0.20 &   0.11 &                             18.61 &   11.92 \\
LinearModel             &                  1 &  35 &  19.95 &   22.0 &                           0.64 &   0.71 &                          0.22 &   0.15 &                              0.04 &    0.03 \\
MLP-rtdl (default)      &                  1 &  35 &  20.11 &   21.0 &                           0.63 &   0.78 &                          0.20 &   0.13 &                             13.07 &    6.33 \\
DANet (default)         &                  1 &  33 &  20.19 &   22.0 &                           0.71 &   0.74 &                          0.21 &   0.15 &                             45.04 &   38.53 \\
TabNet                  &                  1 &  35 &  20.56 &   22.5 &                           0.68 &   0.78 &                          0.25 &   0.18 &                             35.09 &   30.83 \\
DecisionTree (default)  &                  1 &  35 &  21.88 &   23.0 &                           0.63 &   0.70 &                          0.24 &   0.16 &                              0.02 &    0.01 \\
KNN                     &                  1 &  35 &  22.88 &   25.0 &                           0.59 &   0.62 &                          0.20 &   0.15 &                              0.01 &    0.00 \\
TabNet (default)        &                  1 &  35 &  23.52 &   25.0 &                           0.60 &   0.70 &                          0.29 &   0.22 &                             28.39 &   25.93 \\
MLP (default)           &                  1 &  35 &  23.78 &   27.0 &                           0.56 &   0.59 &                          0.24 &   0.18 &                             17.37 &   11.30 \\
KNN (default)           &                  1 &  35 &  24.61 &   26.5 &                           0.54 &   0.57 &                          0.21 &   0.16 &                              0.01 &    0.00 \\
VIME                    &                  3 &  34 &  25.16 &   27.0 &                           0.53 &   0.59 &                          0.17 &   0.13 &                             17.10 &   14.97 \\
STG (default)           &                  1 &  35 &  26.17 &   30.0 &                           0.44 &   0.38 &                          0.18 &   0.11 &                             16.39 &   13.74 \\
VIME (default)          &                  6 &  35 &  30.78 &   33.0 &                           0.23 &   0.05 &                          0.23 &   0.14 &                             15.75 &   14.10 \\
\bottomrule
\end{tabular}
}
\end{table}

Now, we present a table similar to \cref{tab:main-performance}, but with the full set of \ndatasets{} datasets. 
See \cref{tab:ranks-accuracy-all-datasets-all-algs}.
We also include partial results for NAM \citep{agarwal2021neural}, DeepFM \citep{guo2017deepfm}, and TabTransformer \citep{huang2020tabtransformer}.
An important caveat to \cref{tab:ranks-accuracy-all-datasets-all-algs} is that not all algorithms completed on all \ndatasets{} (especially NAM, DeepFM, and TabTransformer) due to memory or timeout issues with our experimental setup described in \cref{sec:analysis}, so these are only rough comparisons.
However, we note that the ordering of the algorithms is still nearly the same as in \cref{tab:main-performance}.

\begin{table}
    \centering
{\small
\caption{Performance of 21 algorithms over all \ndatasets{} datasets according to accuracy. 
\textbf{A caveat is that some algorithms did not complete on all \ndatasets{} datasets (especially DeepFM, TabTransformer, and NAM) due to memory or timeout issues with our experimental setup, so this is only a rough comparison} (however, we note that the ordering of the algorithms is still nearly the same as in \cref{tab:main-performance}).
Columns show the rank over all datasets, the average normalized accuracy (Mean Acc.), and the standard deviation of normalized accuracy across folds (Std.\ Acc.). Min/max/mean/median are taken over all datasets. The rightmost column show the number of datasets where each algorithm ran successfully, out of \ndatasets{}.
}
\label{tab:ranks-accuracy-all-datasets-all-algs}
\begin{tabular}{lrrrrrrrrr}
\toprule
{} & \multicolumn{4}{l}{Rank} & \multicolumn{2}{l}{Mean Acc.} & \multicolumn{2}{l}{Std.\ Acc.} & Num.\ Datasets \\
Algorithm &                min & max &   mean & med.\ &                           mean & med.\ &                          mean & med.\ &                          \\
\midrule
CatBoost       &                  1 &  18 &   5.21 &    4.0 &                           0.88 &   0.94 &                          0.22 &   0.11 &   165 \\
XGBoost        &                  1 &  19 &   5.60 &    4.0 &                           0.88 &   0.96 &                          0.24 &   0.12 &   174 \\
ResNet         &                  1 &  20 &   6.80 &    7.0 &                           0.79 &   0.88 &                          0.22 &   0.10 &   174 \\
LightGBM       &                  1 &  21 &   6.98 &    6.0 &                           0.83 &   0.92 &                          0.27 &   0.12 &   165 \\
SAINT          &                  1 &  20 &   7.55 &    7.0 &                           0.76 &   0.88 &                          0.25 &   0.13 &   138 \\
NODE           &                  1 &  20 &   7.57 &    7.0 &                           0.76 &   0.84 &                          0.23 &   0.14 &   141 \\
RandomForest   &                  1 &  20 &   8.10 &    8.0 &                           0.77 &   0.84 &                          0.22 &   0.10 &   173 \\
FTTransformer  &                  1 &  17 &   8.15 &    8.0 &                           0.76 &   0.81 &                          0.25 &   0.14 &   148 \\
SVM            &                  1 &  19 &   8.41 &    8.0 &                           0.74 &   0.85 &                          0.21 &   0.14 &   143 \\
DANet          &                  1 &  20 &   8.73 &    8.0 &                           0.77 &   0.83 &                          0.26 &   0.15 &   147 \\
MLP-rtdl       &                  1 &  19 &   9.57 &   10.0 &                           0.67 &   0.75 &                          0.20 &   0.09 &   176 \\
DeepFM         &                  1 &  21 &  10.89 &   11.5 &                           0.64 &   0.74 &                          0.26 &   0.19 &    90 \\
TabNet         &                  1 &  21 &  11.05 &   10.0 &                           0.64 &   0.75 &                          0.29 &   0.13 &   168 \\
MLP            &                  1 &  21 &  11.36 &   12.0 &                           0.62 &   0.64 &                          0.21 &   0.12 &   175 \\
DecisionTree   &                  1 &  21 &  11.40 &   12.0 &                           0.60 &   0.65 &                          0.25 &   0.13 &   175 \\
TabTransformer &                  1 &  21 &  11.52 &   12.0 &                           0.58 &   0.66 &                          0.16 &   0.10 &   124 \\
STG            &                  1 &  21 &  11.55 &   12.0 &                           0.60 &   0.69 &                          0.22 &   0.11 &   164 \\
LinearModel    &                  1 &  21 &  12.30 &   14.0 &                           0.51 &   0.57 &                          0.23 &   0.13 &   168 \\
KNN            &                  1 &  21 &  12.83 &   14.0 &                           0.53 &   0.58 &                          0.22 &   0.13 &   167 \\
VIME           &                  1 &  21 &  14.58 &   16.0 &                           0.41 &   0.40 &                          0.21 &   0.12 &   163 \\
NAM            &                  1 &  21 &  15.94 &   17.0 &                           0.35 &   0.26 &                          0.25 &   0.17 &    80 \\
\bottomrule
\end{tabular}
}
\end{table}

\subsubsection{Training time analysis}\label{app:runtime}

In this section we analyze the relative training time required by each algorithm. 
Here we only consider algorithms with their \emph{default} hyperparameters, so no tuning is used.
\cref{tab:runtime-rank} shows a ranking of all algorithms, according to the average training time per 1\,000 training samples.
Similar to \cref{tab:ranks-accuracy-all-datasets-all-algs}, an important caveat to \cref{tab:runtime-rank} is that not all algorithms completed on all \ndatasets{} (especially NAM, DeepFM, and TabTransformer) due to memory or timeout issues with our experimental setup, so these are only rough comparisons.
We also separately present results for datasets of size less than or equal to 1250 in \cref{tab:runtime-rank-tabpfn} (and this table has no such caveat).
%

\begin{table}[t]
    \centering
    \caption{Ranking of all algorithms over all \ndatasets{} datasets, according to average training time per 1\,000 samples. 
    \textbf{A caveat is that some algorithms did not complete on all \ndatasets{} datasets (especially DeepFM, TabTransformer, and NAM) due to memory or timeout issues with our experimental setup, so this is only a rough comparison.}
    Lower ranks indicate lower training times. Rank columns show min, max, and mean ranks over all datasets. Right columns show average training time per 1\,000 samples over all 10 training folds, and the number of datasets considered for each algorithm.
    \label{tab:runtime-rank}}
\begin{tabular}{llllll}
\toprule
{}    & \multicolumn{3}{c}{Rank} & Mean Train Time & Num.\ Datasets \\
Algorithm &                \emph{min} & \emph{max} &   \emph{mean} &  (s/1000 samples) &\\
\midrule
KNN            &                        1 &   5 &   1.66 &                0.04 &   167 \\
DecisionTree   &                        1 &   8 &   2.33 &                0.18 &   175 \\
LinearModel    &                        1 &   4 &   2.45 &                0.05 &   168 \\
RandomForest   &                        1 &   8 &   5.05 &                0.47 &   173 \\
XGBoost        &                        1 &  16 &   5.70 &                1.68 &   174 \\
LightGBM       &                        4 &  14 &   6.67 &                2.64 &   165 \\
SVM            &                        3 &  20 &   6.74 &                3.76 &   141 \\
CatBoost       &                        1 &  19 &   7.32 &               15.36 &   165 \\
MLP-rtdl       &                        2 &  17 &   9.23 &                9.21 &   176 \\
DeepFM         &                        7 &  16 &   9.64 &                5.52 &    90 \\
MLP            &                        3 &  17 &  10.43 &               12.19 &   175 \\
ResNet         &                        4 &  16 &  10.46 &               11.58 &   174 \\
TabTransformer &                        5 &  19 &  12.79 &               17.28 &   122 \\
VIME           &                        8 &  19 &  13.12 &               18.25 &   156 \\
STG            &                        8 &  20 &  13.18 &               15.26 &   164 \\
FTTransformer  &                        7 &  18 &  13.47 &               21.97 &   148 \\
TabNet         &                        9 &  20 &  15.37 &               26.41 &   160 \\
DANet          &                       11 &  21 &  17.20 &               42.10 &   146 \\
NODE           &                        7 &  21 &  17.44 &               60.76 &   141 \\
NAM            &                       12 &  21 &  18.34 &              129.18 &    79 \\
SAINT          &                       10 &  21 &  18.38 &              119.49 &   124 \\
\bottomrule
\end{tabular}
\end{table}

\begin{table}[t]
    \centering
    \caption{Ranking of all 22 algorithms over the 57 datasets of size less than or equal to 1250, according to average training time per 1\,000 samples. Lower ranks indicate lower training times. Rank columns show min, max, and mean ranks over all datasets. Right columns show average training time per 1\,000 samples over all 10 training folds, and the number of datasets considered for each algorithm.
    \label{tab:runtime-rank-tabpfn}}
\begin{tabular}{llllll}
\toprule
{}    & \multicolumn{3}{c}{Rank} & Mean Train Time & \# Datasets \\
Algorithm &                \emph{min} & \emph{max} &   \emph{mean} &  (s/1000 samples) & Num.\ Datasets \\
\midrule
TabPFN        &                        1 &   3 &   1.16 &                0.00 &    57 \\
KNN           &                        1 &   3 &   2.00 &                0.00 &    57 \\
DecisionTree  &                        1 &   4 &   2.88 &                0.02 &    57 \\
LinearModel             &                        3 &   5 &   4.25 &                0.06 &    57 \\
SVM           &                        4 &   8 &   5.23 &                0.22 &    57 \\
LightGBM      &                        5 &  12 &   6.47 &                0.85 &    57 \\
RandomForest  &                        5 &   9 &   6.84 &                0.70 &    57 \\
XGBoost       &                        6 &   9 &   7.35 &                1.25 &    57 \\
CatBoost      &                        8 &  19 &   9.67 &               17.94 &    57 \\
MLP-rtdl      &                        9 &  17 &  11.51 &               20.26 &    57 \\
ResNet        &                       10 &  16 &  12.14 &               22.70 &    57 \\
VIME          &                        9 &  16 &  12.49 &               18.03 &    57 \\
STG           &                        9 &  15 &  12.56 &               18.99 &    57 \\
MLP           &                        9 &  17 &  13.26 &               26.30 &    57 \\
FTTransformer &                       11 &  18 &  13.86 &               29.09 &    57 \\
TabNet        &                       10 &  18 &  15.04 &               32.14 &    57 \\
DANet         &                       15 &  19 &  17.39 &               53.91 &    57 \\
NODE          &                       13 &  19 &  17.40 &               61.94 &    57 \\
SAINT         &                       16 &  19 &  18.51 &              155.07 &    57 \\
\bottomrule
\end{tabular}
\end{table}

\input{new_tabpfn_results}

\subsubsection{HPO Plot} \label{app:hpo}
\begin{figure}
    \centering
    \includegraphics[width=0.7\textwidth]{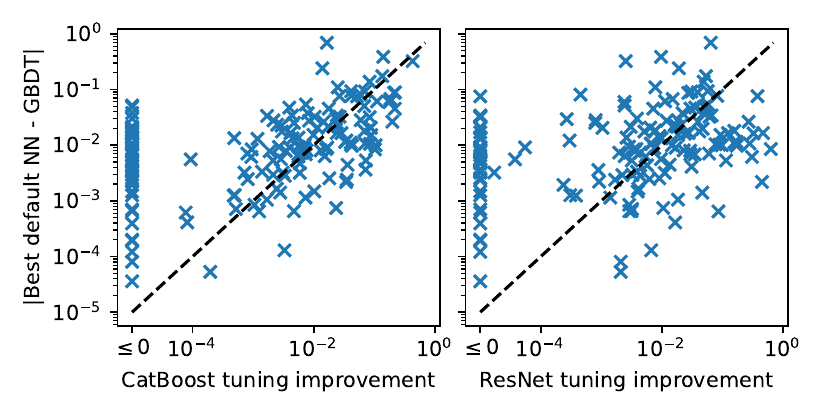}
    \caption{The performance improvement of hyperparameter tuning (horizontal axis) for CatBoost (left) and ResNet (right) compared to the absolute performance difference between the best neural net and the best GBDT, using default hyperparameters (vertical axis). Each point indicates a different dataset, and all values are in normalized log loss. 
    Points on the dotted line indicate that the performance improvement due to hyperparameter tuning is identical to the difference due to NN-vs-GBDT algorithm selection.}
    \label{fig:hparam-tuning-scatter}
\end{figure}

In \cref{fig:venn}, we plotted the performance improvement of hyperparameter tuning on CatBoost, compared to the absolute performance difference between the best neural net and the best GBDT using default hyperparameters. 
Now, we also give the same plot for ResNet. See \cref{fig:hparam-tuning-scatter}.

\subsubsection{Critical difference diagrams}

\begin{figure}
    \centering
    \includegraphics[width=0.8\textwidth]{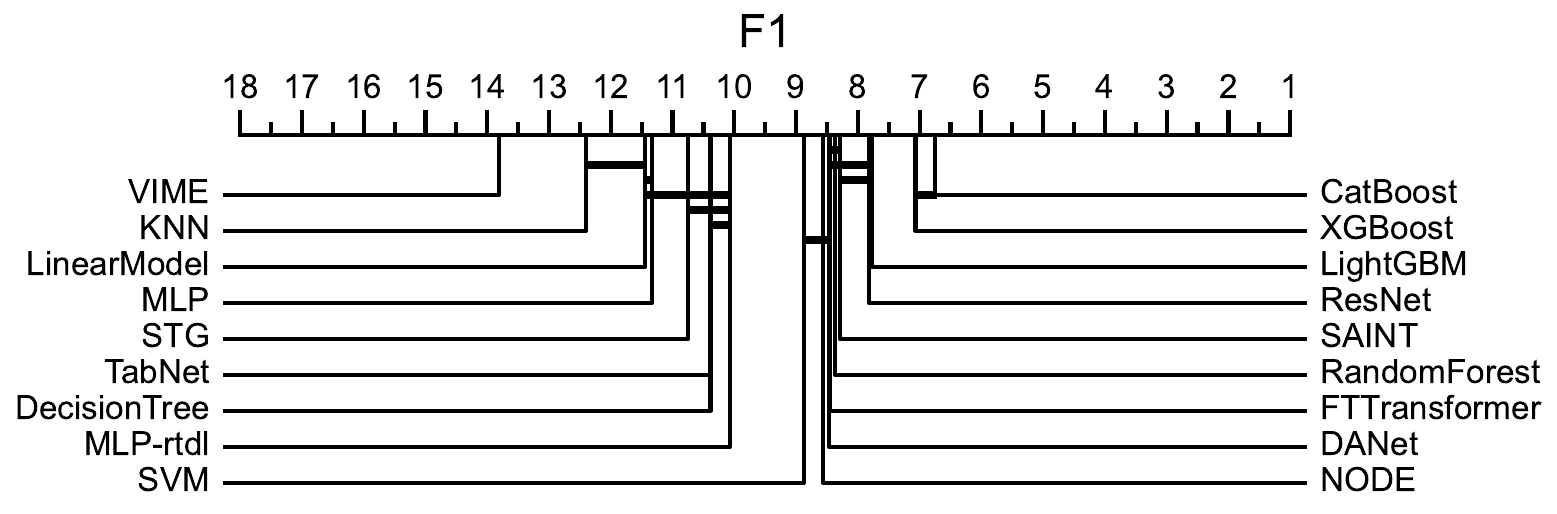}
    \caption{Critical difference diagram comparing all algorithms according to F1 score. Each algorithm's average rank is shown as a horizontal line on the axis. Algorithms which are \emph{not significantly different} are connected by a horizontal black bar.}
    \label{fig:cd-all-algs-f1}
\end{figure}

\cref{fig:cd-all-algs-f1} shows a critical difference plot according to F1 score.
Note that we repeat the Friedman test four times with rankings of the same datasets and algorithms. 
However it is unlikely that our findings would change, given that p-values for these tests without correction are extremely small (p$<10^{-20}$).

\begin{figure}
    \centering
    \includegraphics[width=0.8\textwidth]{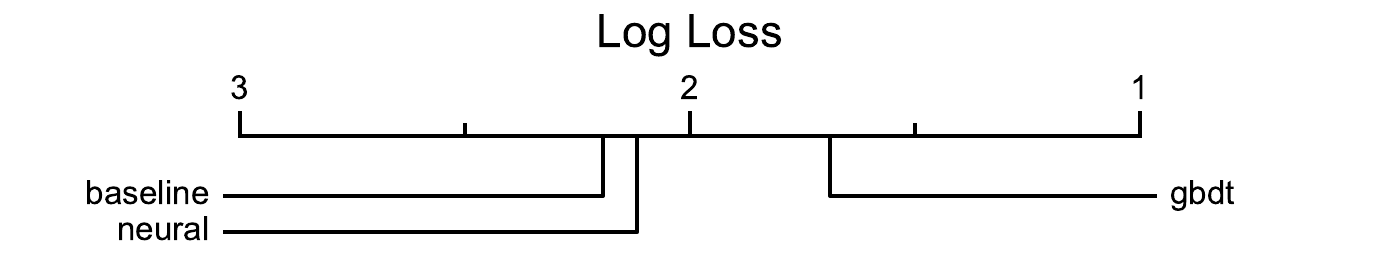}
    \caption{Critical difference plot comparing three algorithm types, according to log loss. Each algorithm's average rank is shown as a horizontal line on the axis. Algorithms which are \emph{not significantly different} are connected by a horizontal black bar.}
    \label{fig:cd-algtype-ll}
\end{figure}

\begin{figure}
    \centering
    \includegraphics[width=0.8\textwidth]{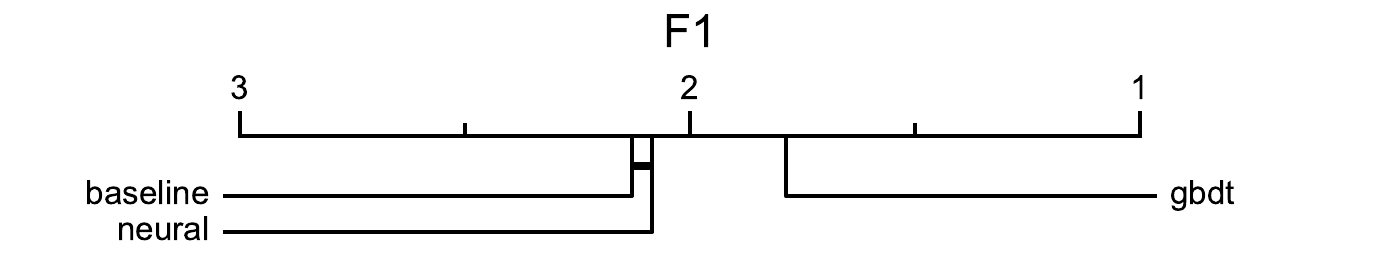}
    \caption{Critical difference plot comparing three algorithm types, according to F1 score. Each algorithm's average rank is shown as a horizontal line on the axis. Algorithms which are \emph{not significantly different} are connected by a horizontal black bar.}
    \label{fig:cd-algtype-f1}
\end{figure}

We compare the performance of each algorithm \emph{family} (GBDTs, NNs, and baselines). 
Here we use all \ndatasets{} datasets. 
We use the same methodology here as in previous sections.
See \cref{fig:cd-algtype-ll} (log loss) and \cref{fig:cd-algtype-f1} (F1 score).

\subsubsection{Venn Diagrams}

\begin{figure}
    \centering
    \includegraphics[width=0.48\textwidth]{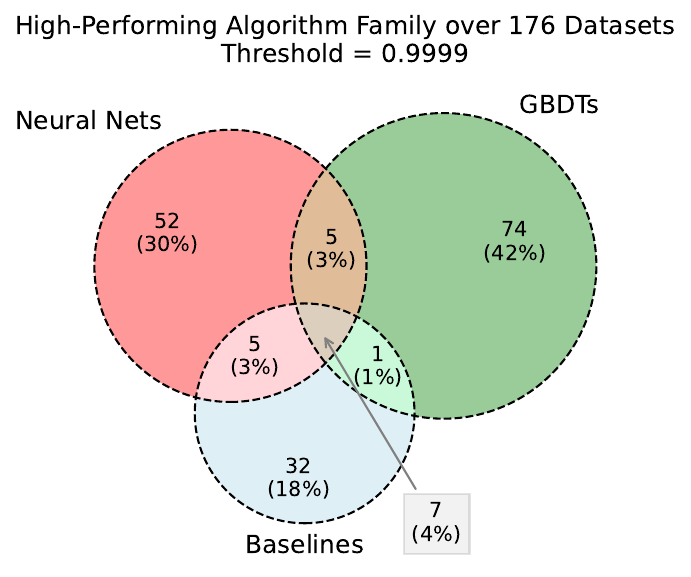}
    \caption{Venn diagram of the number of datasets where an algorithm family is `high-performing', over all \ndatasets{} datasets.
    An algorithm is high-performing if its test accuracy after 0-1 scaling is above a certain threshold.
    While \cref{fig:venn} used a threshold 0.99, this figure uses threshold 0.9999.
    }
    \label{fig:venn_full}
\end{figure}

Recall from \cref{sec:analysis} that for our Venn diagram plots, we split the \nalgs{} algorithms into three \emph{families}: GBDTs (CatBoost, XGBoost, LightGBM), NNs (listed in \cref{sec:analysis}), and baselines (Decision Tree, KNN, LinearModel, RandomForest, SVM).
To compare algorithm performance across datasets, we use min-max scaling as described in \cref{sec:analysis}. 
In \cref{fig:venn}, we said that an algorithm is `high-performing' if it achieves a scaled test-set accuracy of at least $0.99$, and then we determine which algorithm families (GBDTs, NNs, baselines) have a high-performing algorithm. 
Now we compute the same Venn diagram, when tightening the definition of high-performing to $0.9999$ scaled accuracy. See \cref{fig:venn_full}.
In this case, GBDTs are the sole high-performing algorithm family for 42\% of datasets, while NNs are the sole high-performing algorithm family for 30\% of datasets.
However, since these differences are smaller than 0.1\%, they may not be significant to practitioners.

\subsubsection{Dataset size analysis\label{app:dataset-size}}

\begin{figure}
    \centering
    \includegraphics[width=0.9\linewidth]{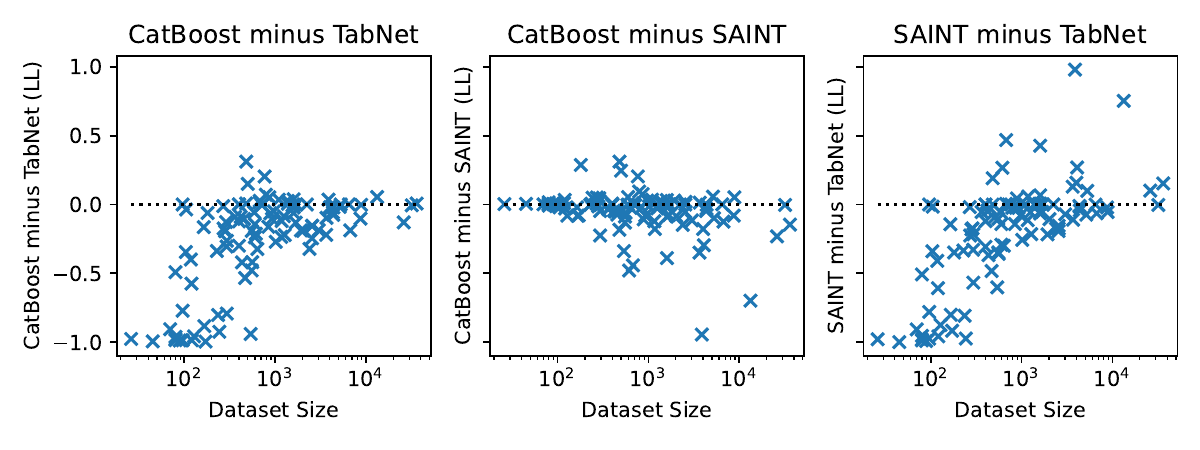}
    \caption{Difference in normalized log loss (lower is better), between three algorithms: CatBoost, SAINT, and TabNet, plotted with dataset size. Points above the dotted line indicate that the second algorithm has lower log loss, meaning better performance than the first.}
    \label{fig:app-size}
\end{figure}

In this section, we investigate the association between dataset size and performance.
In \cref{subsec:metafeature} we show that, when compared to NNs and baselines, GBDTs perform relatively better with larger datasets; this is based on a negative correlations between normalized log loss and dataset size.
However, it is more informative to compare performance of individual algorithms rather than algorithm families.
For example, \cref{fig:app-size} compares the rank of three algorithms: CatBoost, SAINT, and TabNet, with dataset size.
In the left panel (CatBoost minus TabNet), CatBoost outperforms TabNet for all datasets up to size roughly 1\,500. For larger datasets there is little difference between CatBoost and TabNet; this indicates that CatBoost should be chosen over TabNet for smaller datasets, and that both algorithms are comparable for larger datasets 
On the other hand, the center panel (CatBoost minus SAINT), indicates that both CatBoost and SAINT have comparable performance for all datasets \emph{up to} those with size 1\,500; for larger datasets, CatBoost outperforms SAINT. This indicates that CatBoost should be chosen over SAINT for very large datasets, but for small datasets both algorithms are comparable.

\textbf{The main takeaway} from these findings is that practitioners should not focus on choosing an algorithm family, such as NNs or GBDTs, to focus on. For example, TabPFN and TabNet are both neural nets, but TabPFN does comparatively better on smaller datasets, while TabNet does comparatively better on larger datasets.
Rather they can consult our metadataset of results to decide which algorithm is appropriate for their specific use case. General trends are helpful, but not sufficient, for selecting an effective algorithm.

\subsection{Additional experiments from \cref{subsec:metafeature}} \label{app:metafeature}

\begin{figure}
    \centering
    \includegraphics[width=0.8\linewidth]{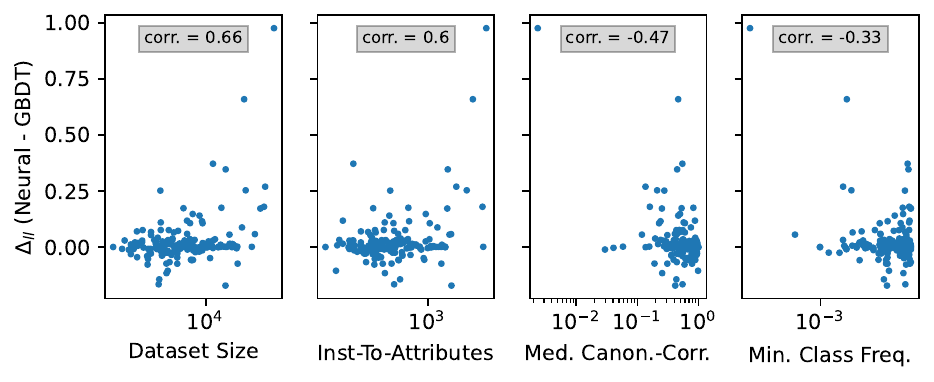}
    \caption{Difference in normalized log loss between the best NN and GBDT (\dll{}) by four dataset properties, for all 10 splits of all \ndatasets{} datasets. 
    Larger correlations mean that a larger value of the property corresponds to larger log loss (worse performance) for the best neural net compared to the best GBDT. All dataset properties are plotted on a log scale.
    }
    \label{fig:corr-nn-gbdt-diff-new}
\end{figure}

\begin{table}[t]
\small
    \centering
    \caption{Selected metafeatures with the largest absolute correlation with the difference in normalized log loss between the best NN and GBDT (\dll{}), over all all \ndatasets{} datasets. Higher correlation values indicate that \emph{larger} values of the metafeature are associated with \emph{worse} NN performance and \emph{stronger} GBDT performance.\label{tab:nn-gbdt-corr-new}}
        \begin{tabular}{lc}
    \toprule
\textbf{Description} & \textbf{Corr. with \dll{}} \\ \midrule
Log number of instances. & 0.63 \\\specialrule{0.2pt}{2pt}{2pt}
Ratio of the size of the dataset to the number of features. & 0.55 \\\specialrule{0.2pt}{2pt}{2pt}
Log of the median canonical correlation between each feature and the target. &  -0.41 \\\specialrule{0.2pt}{2pt}{2pt}
Log of the min.\ target class frequency. &  -0.35 \\
\bottomrule
    \end{tabular}   
\end{table}

Recall from \cref{subsec:metafeature} that \dll{} denotes the difference in normalized log loss between the best neural net and the best GBDT method.
\cref{tab:nn-gbdt-corr-new} and \cref{fig:corr-nn-gbdt-diff-new} shows the dataset properties with the largest absolute correlation with \dll{}.

\begin{figure*}
     \centering
     \includegraphics[width=0.99\linewidth]{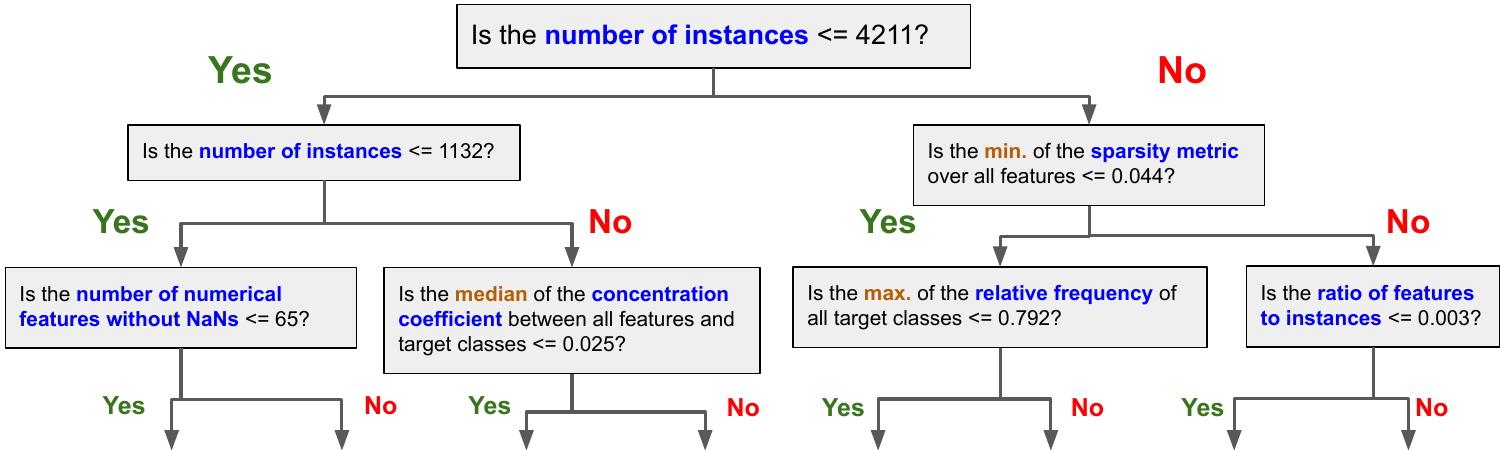}%
     \\
     \includegraphics[width=0.99\linewidth]{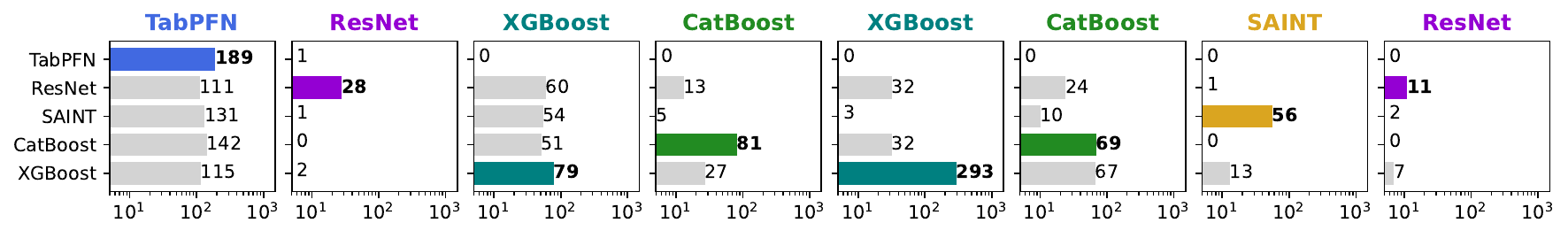}
      \caption{
      Decision tree for picking the best algorithm, based on our experiments across \ndatasets{} datasets.
      The decision tree chooses among the five best-performing algorithms from our experiments: ResNet, SAINT, TabPFN, CatBoost, and XGBoost. 
      Note that the decision splits are based purely on maximizing information gain at that point in the tree, across \ndatasets{} datasets.
      IQR denotes interquartile range, and CanCor \citep{cancor} denotes canonical correlation.}
      \label{fig:alg-decision-tree}
 \end{figure*}

To evaluate the predictive power of dataset properties, we train several decision tree models using the train/test procedure above, with a binary outcome: 1 if \dll$>0$ (the best neural net beats the best GBDT), and 0 otherwise.
\cref{tab:tree-neural-beats-gbdt} shows the performance accuracy of decision trees trained on this task, with varying depth levels; we also include an XGBoost model for comparison.
Finally, we include a visual depiction of a simple depth-3 decision tree, in \cref{fig:alg-decision-tree}.
The decision tree classifies which of the top five algorithms performs the best.
Note that the decision splits are based purely on maximizing information gain at that point in the tree.

\begin{table}[t]
    \centering
    \caption{The test accuracy of tree models for predicting whether the best neural network will outperform the best GBDT model on a tabular dataset. Results are aggregated over \ndatasets{} train/test splits with one dataset family held out for testing in each split. The number of dataset properties used by any model of each model type is listed in the leftmost column. Top rows include decision trees (DT$\leq n$) with maximum depth $n$; the bottom row is an XGBoost model for comparison.
    \label{tab:tree-neural-beats-gbdt}}    
    \begin{tabular}{ccp{2.5cm}}\toprule
        Model & \begin{tabular}{l}Test Accuracy \\ (mean $\pm$ stddev)\end{tabular} & \begin{tabular}{l}Num.\\ Metafeatures \end{tabular}  \\ \midrule
DT$=1$ & 0.54 $\pm$ 0.28 & 3 \\
DT$\leq3$ & 0.60 $\pm$ 0.29 & 21 \\
DT$\leq5$ & 0.59 $\pm$ 0.28 & 183 \\
DT$\leq7$ & 0.64 $\pm$ 0.29 & 382 \\
DT$\leq9$ & 0.66 $\pm$ 0.27 & 423 \\
DT$\leq11$ & 0.66 $\pm$ 0.30 & 529 \\
DT$\leq13$ & 0.65 $\pm$ 0.29 & 561 \\
DT$\leq15$ & 0.68 $\pm$ 0.30 & 594 \\
DT$\leq\infty$ & 0.67 $\pm$ 0.30 & 685 \\
XGBoost & 0.74 $\pm$ 0.33 & 675 \\
         \bottomrule
    \end{tabular}
\end{table}

Finally, we present the metafeatures most-correlated with the difference in log loss between pairs of algorithms. We consider the two best-performing algorithms from each family: CatBoost, XGBoost, ResNet, and SAINT.
See \cref{tab:feature-corrs-ll} and \cref{tab:feature-corrs-ll-general} for statistical and general metafeatures, respectively.


\subsection{Experiments on Regression Datasets}\label{app:regression}

While our experiments focus on classification datasets, the TabZilla codebase is also equipped to handle regression datasets---which have a continuous target variable rather than categorical or binary.
We run experiments using 12 algorithms with 17 tabular regression datasets, using the same experiment design and parameters described in \cref{sec:analysis}. 
Each algorithm is tuned for each dataset by maximizing the R-squared (R2) metric.
The regression datasets used in these experiments have been used in recent studies of machine learning with tabular data~\cite{schafl2022hopular,Popov2020Neural,huang2020tabtransformer,gorishniy2021revisiting}; each dataset corresponds to an OpenML task, and can be preprocessed exactly like the classification datasets used in other experiments. The datasets used in these experiments are ``Bank-Note-Authentication-UCI'' (OpenML task 361002), ``EgyptianSkulls'' (5040), ``Wine'' (190420), ``Wisconsin-breast-cancer-cytology-features'' (361003), ``bodyfat'' (5514), ``california'' (361089), ``chscase-foot'' (5012), ``cleveland'' (2285), ``colleges'' (359942), ``cpu-small'' (4883), ``dataset-sales'' (190418), ``kin8nm'' (2280), ``liver-disorders'' (52948), ``meta'' (4729), ``mv'' (4774), ``pbc'' (4850), and ``veteran'' (4828).

\cref{tab:regression-rankings} shows the rankings of 12 algorithms on these 17 regression datasets, according to the R2 metric calculated on the test set.
The general conclusions are similar to our findings with classification datasets: most algorithms perform well and poorly on \emph{at least one} dataset; however, GBDTs perform particularly well, especially CatBoost.

\begin{table}[t]
\centering
\caption{Performance of 12 algorithms across 17 tabular regression datasets. Columns show the rank over all datasets, the average normalized R2 (Mean R2), and the std.\ dev.\ of normalized R2 across folds (Std.\ R2). Min/max/mean/median of these quantities are taken over all datasets.}
\label{tab:regression-rankings}
\begin{tabular}{lrrrrrrrr}
\toprule
{} & \multicolumn{4}{l}{Rank} & \multicolumn{2}{l}{Mean R2} & \multicolumn{2}{l}{Std. R2} \\
Algorithm &               min & max &   mean & med.\ &                           mean & med.\ &                          mean & med.\ \\
\midrule
CatBoost & 1.0 & 7.0 & 3.0 & 3.0 & 0.96 & 0.98 & 0.11 & 0.02 \\
LightGBM & 1.0 & 11.0 & 4.35 & 3.0 & 0.93 & 0.97 & 0.15 & 0.03 \\
RandomForest & 2.0 & 10.0 & 4.94 & 4.0 & 0.89 & 0.94 & 0.17 & 0.03 \\
XGBoost & 1.0 & 10.0 & 5.29 & 6.0 & 0.89 & 0.99 & 0.18 & 0.03 \\
STG & 1.0 & 11.0 & 6.24 & 7.0 & 0.82 & 0.89 & 0.10 & 0.02 \\
LinearModel & 1.0 & 12.0 & 6.71 & 7.0 & 0.75 & 0.89 & 0.21 & 0.04 \\
MLP & 1.0 & 12.0 & 6.82 & 7.0 & 0.78 & 0.95 & 0.23 & 0.04 \\
NODE & 1.0 & 12.0 & 7.56 & 8.5 & 0.47 & 0.5 & 0.16 & 0.16 \\
TabNet & 3.0 & 12.0 & 7.65 & 7.0 & 0.68 & 0.9 & 0.48 & 0.04 \\
DecisionTree & 2.0 & 12.0 & 7.94 & 8.0 & 0.7 & 0.84 & 0.30 & 0.07 \\
KNN & 2.0 & 12.0 & 8.06 & 8.0 & 0.64 & 0.88 & 0.11 & 0.03 \\
VIME & 3.0 & 12.0 & 9.18 & 10.0 & 0.62 & 0.77 & 0.15 & 0.07 \\
\bottomrule
\end{tabular}
\end{table}



\subsection{Additional Results with Quantile Scaling}

In this section, we discuss dataset preprocessing.
Different papers use a variety of different preprocessing methods, and there is also a wide range in the amount of `built-in' preprocessing techniques inside the algorithms themselves. Therefore, our main results minimize confounding factors by having a consistent, lightweight preprocessing (imputing NaN values). In this section, we compare 13 algorithms on the tabzilla benchmark suite with and without quantile scaling, one of the most popular techniques, for all continuous features.
We use QuantileTransformer from scikit-learn \citep{pedregosa2011scikit}.
We use the same computational setup and experiment design as in our main experiments.
See \cref{tab:benchmark-suite-quantile-scaling}. 
We find that quantile scaling improves the simple algorithms: decision tree, MLP, random forest, and SVM, while it has little effect on the high-performing algorithms.

\begin{table}
\centering
\caption{Performance of 13 algorithms on 36 datasets in the hard dataset benchmark suite, with and without applying quantile scaling to each continuous feature. 
Algorithms with the suffix ``-QSCALE'' use quantile scaling, while those without this suffix use the raw continuous features. Algorithms are ranked according to normalized log loss, and columns show the rank, normalized log loss, and training time, similar to \cref{tab:main-performance}.
\textbf{A caveat is that some algorithms did not complete on all 36 datasets due to memory or timeout issues with our experimental setup, so it is only a rough comparison \emph{among} algorithms} (comparisons between X and X-QSCALE for all X, have the same number of datasets).
\label{tab:benchmark-suite-quantile-scaling}}
\resizebox{\linewidth}{!}{
\begin{tabular}{lrrrrrrrrrr}
\toprule
{} & {} & \multicolumn{4}{l}{Rank} & \multicolumn{2}{l}{Mean LL} & \multicolumn{2}{l}{Std. LL} & \multicolumn{1}{l}{Time /1000 inst.} \\
Algorithm &               min & max &   mean & med.\ &                           mean & med.\ &                          mean & med.\ &                              mean &  med.\ \\
\midrule
XGBoost & 1 & 22 & 4.78 & 3 & 0.07 & 0.04 & 0.10 & 0.06 & 2.02 & 0.28 \\
XGBoost-QSCALE & 1 & 24 & 4.83 & 3 & 0.07 & 0.04 & 0.10 & 0.06 & 1.02 & 0.26 \\
CatBoost & 1 & 19 & 5.60 & 4 & 0.09 & 0.06 & 0.11 & 0.06 & 26.46 & 1.15 \\
CatBoost-QSCALE & 1 & 27 & 6.57 & 4 & 0.12 & 0.08 & 0.10 & 0.06 & 2.40 & 1.19 \\
LightGBM & 1 & 24 & 9.53 & 8 & 0.13 & 0.09 & 0.19 & 0.08 & 1.23 & 0.36 \\
ResNet & 1 & 23 & 9.89 & 9 & 0.19 & 0.17 & 0.12 & 0.08 & 8.27 & 5.22 \\
ResNet-QSCALE & 1 & 25 & 10.43 & 9 & 0.22 & 0.17 & 0.13 & 0.09 & 8.63 & 5.74 \\
SAINT & 1 & 28 & 10.52 & 8 & 0.20 & 0.14 & 0.12 & 0.08 & 130.18 & 92.42 \\
DANet & 2 & 23 & 11.63 & 12 & 0.18 & 0.15 & 0.14 & 0.11 & 58.70 & 52.74 \\
SAINT-QSCALE & 1 & 24 & 11.74 & 10 & 0.23 & 0.15 & 0.09 & 0.05 & 90.53 & 42.05 \\
FTTransformer & 4 & 24 & 12.48 & 11 & 0.25 & 0.20 & 0.13 & 0.11 & 17.41 & 12.64 \\
RandomForest-QSCALE & 3 & 26 & 12.77 & 11 & 0.27 & 0.23 & 0.16 & 0.08 & 0.20 & 0.15 \\
DANet-QSCALE & 2 & 27 & 13.33 & 12 & 0.20 & 0.17 & 0.18 & 0.14 & 59.44 & 57.77 \\
FTTransformer-QSCALE & 5 & 24 & 13.81 & 12 & 0.27 & 0.22 & 0.14 & 0.09 & 19.00 & 15.93 \\
RandomForest & 4 & 27 & 14.46 & 14 & 0.28 & 0.27 & 0.22 & 0.09 & 0.35 & 0.24 \\
MLP-rtdl & 2 & 30 & 14.56 & 12 & 0.36 & 0.25 & 0.16 & 0.13 & 6.33 & 4.21 \\
SVM & 1 & 28 & 15.03 & 14 & 0.29 & 0.21 & 0.14 & 0.08 & 19.73 & 2.81 \\
SVM-QSCALE & 6 & 22 & 15.25 & 17 & 0.21 & 0.17 & 0.13 & 0.12 & 8.85 & 1.91 \\
STG & 1 & 28 & 15.61 & 17 & 0.32 & 0.23 & 0.10 & 0.06 & 15.99 & 15.29 \\
MLP-rtdl-QSCALE & 1 & 29 & 15.92 & 14.5 & 0.39 & 0.31 & 0.18 & 0.14 & 7.35 & 4.85 \\
TabNet & 1 & 30 & 16.88 & 16 & 0.37 & 0.27 & 0.26 & 0.12 & 27.02 & 27.10 \\
NODE & 9 & 27 & 17.50 & 17 & 0.37 & 0.34 & 0.08 & 0.07 & 153.72 & 124.27 \\
NODE-QSCALE & 6 & 26 & 17.76 & 18 & 0.36 & 0.33 & 0.08 & 0.07 & 171.73 & 147.36 \\
MLP & 3 & 29 & 17.81 & 18.5 & 0.43 & 0.40 & 0.15 & 0.14 & 8.86 & 4.36 \\
LinearModel & 4 & 29 & 17.97 & 18.5 & 0.47 & 0.37 & 0.11 & 0.07 & 0.04 & 0.02 \\
LinearModel-QSCALE & 5 & 30 & 18.06 & 17 & 0.47 & 0.38 & 0.12 & 0.08 & 0.01 & 0.01 \\
VIME & 6 & 28 & 21.44 & 24 & 0.49 & 0.48 & 0.09 & 0.08 & 20.79 & 15.18 \\
DecisionTree-QSCALE & 8 & 30 & 21.54 & 23 & 0.60 & 0.59 & 0.36 & 0.20 & 0.05 & 0.01 \\
DecisionTree & 9 & 31 & 21.86 & 23 & 0.62 & 0.62 & 0.38 & 0.20 & 0.11 & 0.01 \\
KNN & 6 & 31 & 23.61 & 25 & 0.68 & 0.70 & 0.35 & 0.21 & 0.03 & 0.00 \\
KNN-QSCALE & 5 & 30 & 23.94 & 25.5 & 0.70 & 0.75 & 0.34 & 0.23 & 0.01 & 0.00 \\
\bottomrule
\end{tabular}
}
\end{table}



\subsection{Experiments with Additional Hyperparameter Optimization}\label{app:more-hpo}

In our main experiments, for hyperparameter optimization (HPO), we ran 30 iterations of random search for all algorithms.
In this section, we test the impact of additional HPO for four algorithms: XGBoost, CatBoost, LightGBM, and RandomForest.
We did not run additional HPO experiments on any neural net methods due to the substantial compute resources required.

For each algorithm, we run 100 iterations of HPO using the default Optuna \citep{akiba2019optuna} algorithm (tree-structured Parzen Estimator), optimizing log loss. 
We run these HPO experiments on all 36 datasets in the TabZilla benchmark suite.
All hyperparameter ranges can be viewed in our repository in the folder \url{https://github.com/naszilla/tabzilla/tree/main/TabZilla/models}.

\cref{tab:more-hpo} shows the performance of these HPO experiments (algorithm suffix ``(HPO)''), compared with with the performance of the default hyperparameters (suffix ``(default)''), and the performance after 30 iterations of random hyperparameter search as in our main results (no suffix). As expected, additional hyperparameter tuning improves the performance of XGBoost, CatBoost, LightGBM, and RandomForest. 

\begin{table}
\centering
\caption{Performance of all algorithms on all 36 datasets in the hard dataset benchmark suite, including the performance after 100 iterations of HPO (algorithms with suffix ``(HPO)''), and the default hyperparameters (algorithms with suffix ``(default)''). Algorithm names without any suffix indicate performance after 30 iterations of random hyperparameter search, as in our main results. Algorithms are ranked according to normalized log loss, and columns show the rank, normalized log loss, and training time, similar to \cref{tab:main-performance}.
\textbf{A caveat is that some algorithms did not complete on all 36 datasets due to memory or timeout issues with our experimental setup, so it is only a rough comparison \emph{among} algorithms} (comparisons between X, X-default, and X-HPO, for all X, have the same number of datasets).
\label{tab:more-hpo}}
\resizebox{\linewidth}{!}{
\begin{tabular}{lrrrrrrrrrr}
\toprule
{} & {} & \multicolumn{4}{l}{Rank} & \multicolumn{2}{l}{Mean LL} & \multicolumn{2}{l}{Std. LL} & \multicolumn{1}{l}{Time /1000 inst.} \\
Algorithm &               min & max &   mean & med.\ &                           mean & med.\ &                          mean & med.\ &                              mean &  med.\ \\
\midrule
XGBoost (HPO) & 1 & 30 & 6.28 & 4 & 0.03 & 0.02 & 0.04 & 0.03 & 6.36 & 1.54 \\
XGBoost & 1 & 28 & 6.83 & 5 & 0.03 & 0.02 & 0.04 & 0.03 & 2.02 & 0.28 \\
CatBoost (HPO) & 1 & 28 & 6.88 & 6 & 0.05 & 0.02 & 0.04 & 0.03 & 16.75 & 1.82 \\
CatBoost & 1 & 20 & 7.23 & 6 & 0.05 & 0.02 & 0.04 & 0.02 & 26.46 & 1.15 \\
LightGBM (HPO) & 1 & 22 & 7.38 & 4.5 & 0.04 & 0.02 & 0.04 & 0.03 & 0.64 & 0.20 \\
XGBoost (default) & 1 & 27 & 9.06 & 8 & 0.07 & 0.03 & 0.03 & 0.03 & 1.76 & 0.41 \\
CatBoost (default) & 1 & 33 & 11.17 & 10 & 0.10 & 0.04 & 0.03 & 0.02 & 29.53 & 0.97 \\
LightGBM & 1 & 30 & 11.94 & 11 & 0.06 & 0.03 & 0.07 & 0.04 & 1.23 & 0.36 \\
ResNet & 1 & 26 & 12.20 & 12 & 0.10 & 0.04 & 0.04 & 0.04 & 8.27 & 5.22 \\
SAINT & 1 & 34 & 13.33 & 11 & 0.08 & 0.05 & 0.04 & 0.03 & 130.18 & 92.42 \\
LightGBM (default) & 2 & 34 & 13.81 & 12 & 0.07 & 0.05 & 0.05 & 0.04 & 1.46 & 0.61 \\
DANet & 2 & 31 & 14.81 & 14 & 0.06 & 0.05 & 0.05 & 0.04 & 58.70 & 52.74 \\
FTTransformer & 2 & 29 & 15.69 & 15 & 0.10 & 0.06 & 0.04 & 0.03 & 17.41 & 12.64 \\
SAINT (default) & 2 & 37 & 16.18 & 13 & 0.08 & 0.05 & 0.04 & 0.04 & 111.07 & 83.68 \\
RandomForest (HPO) & 4 & 35 & 16.29 & 13 & 0.14 & 0.08 & 0.08 & 0.03 & 0.33 & 0.20 \\
ResNet (default) & 2 & 35 & 17.00 & 16 & 0.13 & 0.06 & 0.06 & 0.04 & 7.28 & 4.72 \\
MLP-rtdl & 2 & 37 & 17.92 & 16 & 0.18 & 0.09 & 0.06 & 0.04 & 6.33 & 4.21 \\
RandomForest & 5 & 33 & 18.03 & 16 & 0.15 & 0.10 & 0.08 & 0.03 & 0.35 & 0.24 \\
STG & 1 & 35 & 18.48 & 20 & 0.14 & 0.06 & 0.04 & 0.03 & 15.99 & 15.29 \\
SVM & 1 & 34 & 18.62 & 18 & 0.11 & 0.07 & 0.04 & 0.03 & 19.73 & 2.81 \\
TabNet & 1 & 37 & 20.30 & 18 & 0.14 & 0.12 & 0.09 & 0.06 & 27.02 & 27.10 \\
RandomForest (default) & 4 & 34 & 20.57 & 21 & 0.23 & 0.10 & 0.03 & 0.02 & 0.32 & 0.27 \\
MLP & 3 & 36 & 21.06 & 21.5 & 0.21 & 0.12 & 0.05 & 0.04 & 8.86 & 4.36 \\
FTTransformer (default) & 5 & 36 & 21.17 & 23 & 0.19 & 0.07 & 0.05 & 0.04 & 15.71 & 11.42 \\
NODE & 11 & 32 & 21.27 & 20 & 0.19 & 0.12 & 0.03 & 0.02 & 153.72 & 124.27 \\
MLP-rtdl (default) & 1 & 39 & 21.69 & 21.5 & 0.29 & 0.13 & 0.11 & 0.06 & 5.82 & 3.89 \\
DANet (default) & 6 & 37 & 21.74 & 23 & 0.12 & 0.09 & 0.08 & 0.05 & 40.59 & 38.95 \\
LinearModel & 8 & 37 & 22.09 & 22.5 & 0.26 & 0.13 & 0.04 & 0.03 & 0.04 & 0.02 \\
NODE (default) & 11 & 34 & 22.70 & 23 & 0.21 & 0.14 & 0.03 & 0.02 & 52.26 & 42.19 \\
SVM (default) & 2 & 35 & 23.57 & 26 & 0.15 & 0.08 & 0.04 & 0.02 & 4.19 & 0.80 \\
TabNet (default) & 2 & 38 & 24.94 & 27 & 0.22 & 0.14 & 0.11 & 0.07 & 24.04 & 23.40 \\
MLP (default) & 4 & 39 & 25.28 & 26 & 0.34 & 0.17 & 0.10 & 0.07 & 8.13 & 4.44 \\
VIME & 4 & 34 & 25.91 & 29 & 0.23 & 0.18 & 0.04 & 0.03 & 20.79 & 15.18 \\
STG (default) & 10 & 38 & 26.81 & 27 & 0.25 & 0.14 & 0.03 & 0.02 & 13.72 & 13.20 \\
DecisionTree & 14 & 37 & 27.29 & 28 & 0.29 & 0.21 & 0.13 & 0.08 & 0.11 & 0.01 \\
KNN & 6 & 37 & 29.27 & 31 & 0.28 & 0.26 & 0.12 & 0.09 & 0.03 & 0.00 \\
DecisionTree (default) & 15 & 39 & 31.91 & 35 & 0.60 & 0.67 & 0.30 & 0.15 & 0.12 & 0.02 \\
VIME (default) & 18 & 39 & 32.00 & 33 & 0.38 & 0.31 & 0.09 & 0.02 & 20.10 & 12.79 \\
KNN (default) & 6 & 39 & 33.48 & 36 & 0.59 & 0.55 & 0.22 & 0.17 & 0.03 & 0.00 \\
\bottomrule
\end{tabular}
}
\end{table}



\subsection{Forward Feature Selection for Identifying Important Dataset Attributes}\label{app:ffs-metafeatures}

In this section, we present a different method for determining which dataset attributes are related to performance differences between algorithms. 
Here we use greedy forward feature selection~\cite{gelsema2014pattern} to identify important dataset attributes. 
In these experiments, we study the problem of predicting the difference in normalized log loss between CatBoost and ResNet (two very effective GBDT and NN algorithms), using metafeatures.

At a high level, greedy forward feature selection selects metafeatures sequentially which improve the performance of the meta-model. 
To evaluate performance we use leave-one-dataset-out cross validation: each dataset contributes 10 folds to the overall metadataset, so each fold includes 10 instances for validation and all remaining instances for training.

In these experiments, we first use an XGB regressor fit on the entire metadataset to select 200 features for consideration. 
Then we use greedy forward selection, implemented in the python package mlxtend (\url{https://rasbt.github.io/mlxtend}), using an XGB regressor as a meta-model. 
The first five selected features are (in order):
\begin{enumerate}
    \item Number of features normally-distributed, according to the Shapiro-Wilk test.
    \item Median value of the minimum of all features.
    \item Median value of the sparsity of all features.~\cite{salama2013employment}
    \item Interquartile range of the mean value of all features.
    \item Mean of the harmonic mean of all features.
\end{enumerate}


\begin{table}[t]
    \centering
    {\small
    \caption{Dataset metafeatures that are most correlated with the difference in normalized log loss between pairs of algorithms (the loss of Alg.\ 1 minus Alg.\ 2). Correlations are taken over all 10 splits of all 133 datasets in which CatBoost, XGBoost, ResNet, and SAINT ran successfully. The 10 dataset attributes with the largest absolute correlation are listed for each pair of algorithms. Attribute names correspond to the naming convention used by PyMFE.\label{tab:feature-corrs-ll}}
  \begin{tabular}{llrp{9.5cm}}
\toprule
  Alg. 1 & Alg. 2 &      Corr. &                                       Attribute Name \\
\midrule
CatBoost & ResNet &  -0.25 &                     Maximum skewness of all features.\\ 
CatBoost & ResNet &  -0.24 &                   Range of the skewness of all features.\\ 
CatBoost & ResNet &  -0.23 &                 Log of the standard deviation of the kurtosis of all features. \\
CatBoost & ResNet &  -0.23 &                Log of the standard deviation of the skewness of all features.\\
CatBoost & ResNet &   0.22 &                  Log of the median of the absolute value of the covariance between all feature pairs.\\ 
CatBoost & ResNet &   0.21 &                  Log of the median of the standard deviation of all features. \\
CatBoost & ResNet &   0.21 &            Log of the median of the variance of all features.  \\  
CatBoost & ResNet &   0.20 &                Log of the median of the maximum value of all features.\\ 
CatBoost & ResNet &   0.20 &    Best performance of a naive Bayes classifier trained over 10-fold CV.\\
\midrule
CatBoost &  SAINT &   0.26 &  Best performance over all 10 folds, of 10-fold CV of a single-node decision tree fit using the least-informative feature. \\ 
CatBoost &  SAINT &   0.25 & Average performance over 10-fold CV of a single-node decision tree fit using the least-informative feature. \\ 
CatBoost &  SAINT &   0.24 &   Median performance over 10 folds, for 10-fold CV of a single-node decision tree fit using the least-informative feature.\\ 
CatBoost &  SAINT &  -0.24 &     Log of the worst performance over 10-fold CV of a single-node decision tree fit using the most-informative feature.\\ 
CatBoost &  SAINT &  -0.23 & Log of the kurtosis of the performance of a single-node decision tree fit using the most-informative feature, over 10-fold CV. \\ 
CatBoost &  SAINT &   0.23 &                Log of the Shannon entropy of the target. \\ 
CatBoost &  SAINT &  -0.23 &    Log of the best performance of elite-nearest-neighbor over 10-fold CV. \\ 
\midrule
 XGBoost & ResNet &  -0.29 & Maximum skewness of all features.\\
 XGBoost & ResNet &  -0.28 & Range of the skewness of all features.\\
 XGBoost & ResNet &  -0.28 &            Log of the standard deviation of the kurtosis of all features.\\
 XGBoost & ResNet &  -0.27 &           Log of the standard deviation of the skewness of all features\\
 XGBoost & ResNet &   0.25 &             Log of the median value of the absolute covariance between all pairs of features.\\ 
 XGBoost & ResNet &   0.24 &          Log of the median standard deviation of all features.\\
 XGBoost & ResNet &   0.23 &          Log of the median variance of all features. \\
 XGBoost & ResNet &   0.23 &           Log of the median  maximum-value of all features.\\
 XGBoost & ResNet &   0.22 & Best performance of a naive Bayes classifier over 10-fold CV.\\
 XGBoost & ResNet &   0.22 &          Log of the best performance of a naive Bayes classifier over 10-fold CV.\\
 \midrule
 XGBoost &  SAINT &  -0.23 &  Noisiness of the features: $(\sum_i S_i - \sum_i MI(i, y)) / \sum_i MI(i, y)$, where $S_i$ is the entropy of feature $i$, and $MI(i, y)$ is the mutual information between feature $i$ and the target $y$.\\
 XGBoost &  SAINT &  -0.22 &             Log of the standard deviation of the kurtosis of all features. \\
 XGBoost &  SAINT &  -0.20 &            Log of the standard deviation of the skewness of all features.    \\
 XGBoost &  SAINT &  -0.20 &    Best performance over 10-fold CV of a single-node decision tree fit using the most-informative feature.\\
 XGBoost &  SAINT &  -0.20 &  Maximum skewness of all features.\\
 XGBoost &  SAINT &   0.19 &        Log of the Shannon entropy of the target.       \\
 XGBoost &  SAINT &  -0.19 &  Log of the standard deviation of the absolute correlation between all pairs of features. \\
 XGBoost &  SAINT &  -0.18 &  Range of the skewness of all features.\\
 XGBoost &  SAINT &   0.18 &  Log of the range of the performance of a decision tree trained on a random attribute, over 10-fold CV.\\
\bottomrule
\end{tabular}
}
\end{table}

\begin{table}[t]
    \centering
    {\small
    \caption{Dataset metafeatures that are most correlated with the difference in normalized log loss between pairs of algorithms (the loss of Alg.\ 1 minus Alg.\ 2). Correlations are taken over all 10 splits of all 133 datasets in which CatBoost, XGBoost, ResNet, and SAINT ran successfully. The 10 dataset attributes with the largest absolute correlation are listed for each pair of algorithms. Attribute names correspond to the naming convention used by PyMFE.\label{tab:feature-corrs-ll-general}}
  \begin{tabular}{llrl}
\toprule
  Alg. 1 & Alg. 2 &      Corr. &                                       Attribute Name \\
\midrule
CatBoost & ResNet &  -0.14 &         Log of the size of the dataset.\\
CatBoost & ResNet &   0.14 &    Log of the ratio of number of features to dataset size.\\
CatBoost & ResNet &  -0.14 &    Log of the ratio of dataset size to number of features.\\
CatBoost & ResNet &   0.06 &   Number of numerical features.\\
CatBoost & ResNet &   0.06 & Number of features.\\
CatBoost & ResNet &  -0.06 &Range of the relative frequency of each target class.    \\
CatBoost & ResNet &  -0.06 &   Log of the maximum frequency of any target class.\\
CatBoost & ResNet &  -0.05 &   Interquartile range of the relative frequency of all target classes. \\
CatBoost & ResNet &  -0.04 &    Maximum relative frequency of any target class.  \\\midrule
CatBoost &  SAINT &  -0.19 &  Standard deviation of the relative frequency of all target classes.\\
CatBoost &  SAINT &   0.18 &  Kurtosis of the relative frequency of all target classes.\\
CatBoost &  SAINT &  -0.18 &  Maximum relative frequency of all target classes.   \\
CatBoost &  SAINT &  -0.18 &      Mean relative frequency of all target classes.\\
CatBoost &  SAINT &  -0.17 & Log of the median relative frequency of all target classes.\\
CatBoost &  SAINT &   0.16 &     Number of target classes.   \\
CatBoost &  SAINT &   0.15 & Skewness of the relative frequency of target classes. \\\midrule
 XGBoost & ResNet &  -0.20 &        Log of the dataset size.\\
 XGBoost & ResNet &  -0.20 & Log of the ratio of dataset size to number of features.    \\
 XGBoost & ResNet &   0.20 &     Log of the ratio of number of features to dataset size. \\
 XGBoost & ResNet &   0.11 &  Log of the minimum  relative frequency of all target classes.\\
 XGBoost & ResNet &   0.08 &     Number of numerical features. \\
 XGBoost & ResNet &   0.08 &   Median relative frequency of all target classes.\\
 XGBoost & ResNet &   0.07 &  Minimum relative frequency of all target classes.\\\midrule
 XGBoost &  SAINT &  -0.17 & Standard deviation of the relative frequency of all target classes. \\
 XGBoost &  SAINT &  -0.16 &   Log of the dataset size. \\ 
 XGBoost &  SAINT &  -0.16 &  Interquartile range of the relative frequency of all target classes.  \\
 XGBoost &  SAINT &   0.15 &   Log of the ratio of dataset size to number of features.  \\
 XGBoost &  SAINT &  -0.15 & Log of the ratio of number of features to dataset size.     \\
 XGBoost &  SAINT &  -0.15 & Maximum relative frequency of all target classes.\\
 XGBoost &  SAINT &  -0.13 &   Range of the relative frequency of all target classes.\\
 XGBoost &  SAINT &  -0.13 & Mean of the relative frequency of all target classes. \\
 XGBoost &  SAINT &  -0.12 & Median of the relative frequency of all target classes. \\
\bottomrule
\end{tabular}
}
\end{table}

\section{Additional Details from \cref{sec:tabzilla}} \label{app:tabzilla}

\cref{tab:benchmark-alg-perf} compares the performance of all algorithms on the benchmark suite.
However, an important caveat to \cref{tab:ranks-accuracy-all-datasets-all-algs} is that not all algorithms completed on all 36 datasets due to memory or timeout issues with our experimental setup, as discussed in \cref{subsec:relative-performance}, so these are only rough comparisons.

\begin{table}
\centering
\caption{Performance of algorithms across the Tabular Benchmark Suite. 
\textbf{An important caveat is that not all algorithms completed on all 36 datasets due to memory or timeout issues with our experimental setup, so these are only rough comparisons.}
Columns show the rank over all algorithms, the average normalized log loss (Mean LL), the standard deviation of normalized log loss across folds (Std.\ LL), and the train time per 1000 instances. Min/max/mean/median of these quantities are taken over all completed datasets for each algorithm.}
\label{tab:benchmark-alg-perf}
\resizebox{\linewidth}{!}{
\begin{tabular}{lrrrrrrrrrr}
\toprule
{} & {} & \multicolumn{4}{l}{Rank} & \multicolumn{2}{l}{Mean LL} & \multicolumn{2}{l}{Std. LL} & \multicolumn{1}{l}{Time /1000 inst.} \\
Algorithm &               min & max &   mean & med.\ &                           mean & med.\ &                          mean & med.\ &                              mean &  med.\ \\
\midrule
XGBoost & 1 & 15 & 3.69 & 2 & 0.07 & 0.04 & 0.11 & 0.06 & 2.02 & 0.28 \\
CatBoost & 1 & 13 & 4.14 & 3 & 0.09 & 0.06 & 0.11 & 0.07 & 26.46 & 1.15 \\
TabPFN$^*$ & 1 & 13 & 5.29 & 4 & 0.19 & 0.13 & 0.11 & 0.07 & 0.48 & 0.01 \\
LightGBM & 1 & 15 & 6.62 & 6.5 & 0.14 & 0.09 & 0.20 & 0.09 & 1.23 & 0.36 \\
ResNet & 1 & 14 & 6.77 & 6 & 0.20 & 0.17 & 0.13 & 0.08 & 8.27 & 5.22 \\
SAINT & 1 & 18 & 7.00 & 6 & 0.20 & 0.15 & 0.12 & 0.08 & 130.18 & 92.42 \\
DANet & 2 & 16 & 8.04 & 8 & 0.19 & 0.16 & 0.15 & 0.11 & 58.70 & 52.74 \\
FTTransformer & 2 & 15 & 8.24 & 8 & 0.25 & 0.20 & 0.13 & 0.12 & 17.41 & 12.64 \\
MLP-rtdl & 2 & 19 & 9.42 & 8 & 0.37 & 0.25 & 0.16 & 0.13 & 6.33 & 4.21 \\
RandomForest & 2 & 17 & 9.43 & 9 & 0.29 & 0.28 & 0.23 & 0.09 & 0.35 & 0.24 \\
SVM & 1 & 18 & 10.00 & 10 & 0.30 & 0.22 & 0.14 & 0.08 & 19.73 & 2.81 \\
STG & 1 & 18 & 10.29 & 11 & 0.33 & 0.26 & 0.11 & 0.06 & 15.99 & 15.29 \\
TabNet & 1 & 19 & 11.03 & 10 & 0.39 & 0.29 & 0.28 & 0.11 & 27.02 & 27.10 \\
NODE & 6 & 17 & 11.40 & 11 & 0.38 & 0.35 & 0.08 & 0.07 & 153.72 & 124.27 \\
MLP & 3 & 18 & 11.42 & 11.5 & 0.45 & 0.39 & 0.15 & 0.14 & 8.86 & 4.36 \\
LinearModel & 4 & 19 & 11.68 & 11.5 & 0.48 & 0.39 & 0.12 & 0.07 & 0.04 & 0.02 \\
VIME & 5 & 18 & 13.91 & 15 & 0.51 & 0.48 & 0.09 & 0.08 & 20.79 & 15.18 \\
DecisionTree & 6 & 19 & 14.26 & 16 & 0.63 & 0.62 & 0.39 & 0.22 & 0.11 & 0.01 \\
KNN & 4 & 19 & 15.42 & 17 & 0.71 & 0.74 & 0.36 & 0.21 & 0.03 & 0.00 \\
\bottomrule
\end{tabular}
}
\end{table}

%% file: new_tabpfn_results.tex
\subsection{Additional TabPFN Results}\label{app:tabpfn}

In this section, we show additional results that include two modified versions of TabPFN, one that uses a random subset of 3000 training points (referred to as TabPFN-3k, or TabPFN$^*$ in the main paper), and one that uses 1000 randomly sampled training points (TabPFN-1k). As an ablation, we also test a version of CatBoost that uses a subset of 1000 training points (CatBoost-1k). 

\begin{figure}
    \centering
    \includegraphics[width=0.8\textwidth]{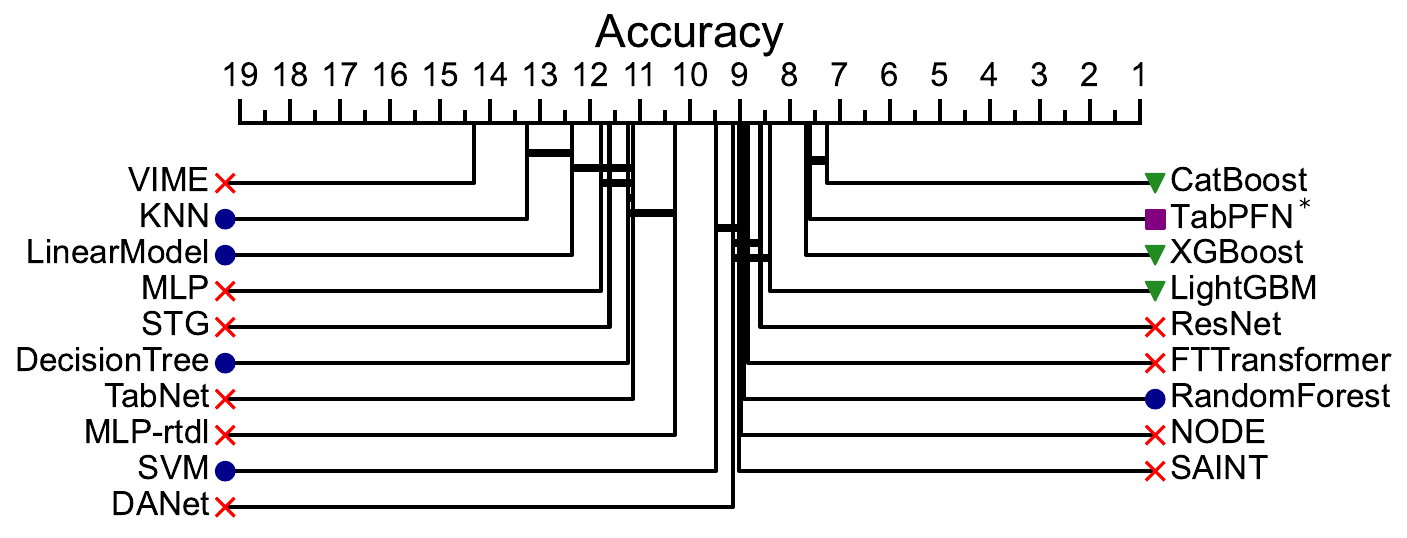}
    \caption{Critical difference plot comparing all algorithms according to their mean accuracy rank over 98 datasets. Each algorithm's average rank is shown as a horizontal line on the axis. Sets of algorithms which are \emph{not significantly different} are connected by a horizontal black bar. }
    \label{fig:cd-tabpfn-acc}
\end{figure}

\begin{figure}
    \centering
    \includegraphics[width=0.9\textwidth]{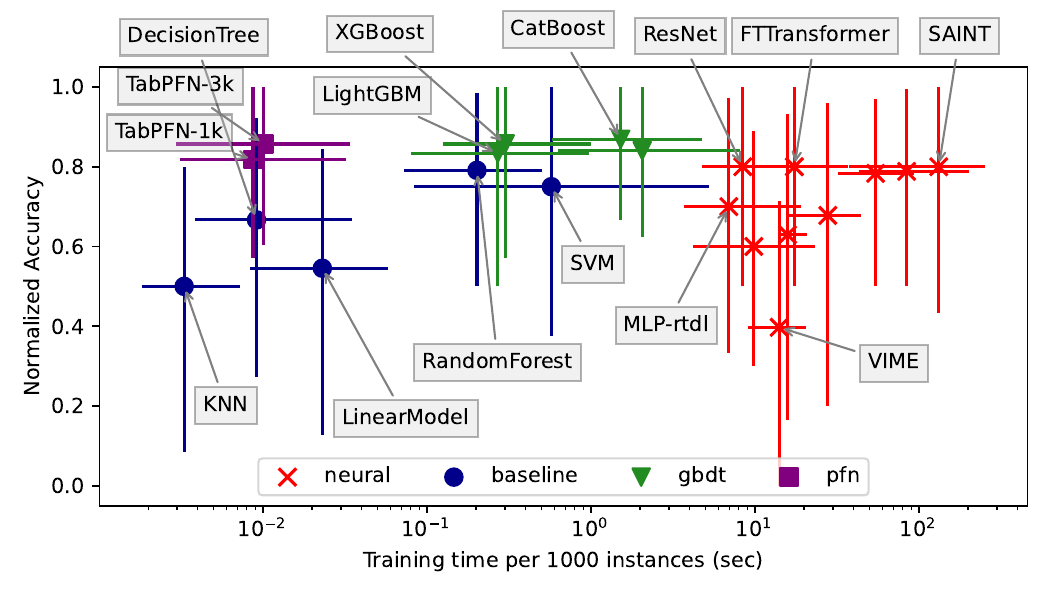}
    \caption{Median runtime vs.\ median normalized accuracy for each algorithm, including two variants of TabPFN, over 98 datasets. The bars span the 20th to 80th percentile over all datasets.}
    \label{fig:accuracy-train-time-with-tabpfn}
\end{figure}

This analysis includes a subset of 98 datasets and 21 algorithms (including TabPFN-3k, TabPFN-1k, and CatBoost-1k), where all algorithms produce a result for all datasets. 
For a table similar to \cref{tab:main-performance}, but with the TabPFN and CatBoost ablations, see \cref{tab:performance-tabpfn-acc} (where the performance metric is accuracy).
See also the same tables for log loss (\cref{tab:performance-tabpfn-ll}) and F1 score (\cref{tab:performance-tabpfn-f1}).
Next, we present critical difference plots similar to \cref{fig:cd-all-algs-ll}, but where the performance metric is accuracy: \cref{fig:cd-tabpfn-acc}.
Finally, we present a plot of accuracy vs.\ time, similar to \cref{fig:runtime-vs-acc-annotations}, but which includes the TabPFN and CatBoost ablations: \cref{fig:accuracy-train-time-with-tabpfn}.

\begin{table}[t]
\centering
\caption{Performance of algorithms across 98 datasets, where the algorithms include two modified versions of TabPFN. Columns show the algorithm family (GBDT, NN, baseline, or PFN), rank over all datasets, the average normalized accuracy (Mean Acc.), the std.\ dev.\ of normalized accuracy across folds (Std.\ Acc.), and the train time in seconds per 1000 instances. Min/max/mean/median of these quantities are taken over all datasets.}
\label{tab:performance-tabpfn-acc}
\resizebox{\linewidth}{!}{
\begin{tabular}{lcrrrrrrrrrr}
\toprule
{} & {} & \multicolumn{4}{l}{Rank} & \multicolumn{2}{l}{Mean Acc.} & \multicolumn{2}{l}{Std. Acc.} & \multicolumn{2}{l}{Time /1000 inst.} \\
Algorithm & Class &               min & max &   mean & med.\ &                           mean & med.\ &                          mean & med.\ &                              mean &  med.\ \\
\midrule
CatBoost & GBDT & 1 & 19 & 6.12 & 5 & 0.87 & 0.93 & 0.3 & 0.22 & 21.7 & 2.08 \\
TabPFN-3k & PFN & 1 & 19 & 6.43 & 5 & 0.84 & 0.92 & 0.26 & 0.19 & 0.25 & 0.01 \\
CatBoost-1k & GBDT & 1 & 21 & 7.02 & 5.5 & 0.85 & 0.9 & 0.3 & 0.23 & 6.76 & 2.4 \\
XGBoost & GBDT & 1 & 19 & 7.85 & 6.5 & 0.81 & 0.89 & 0.33 & 0.22 & 0.81 & 0.37 \\
TabPFN-1k & PFN & 1 & 21 & 7.94 & 7 & 0.8 & 0.91 & 0.27 & 0.19 & 0.25 & 0.01 \\
ResNet & NN & 1 & 21 & 8.73 & 8 & 0.75 & 0.83 & 0.3 & 0.21 & 16.01 & 9.34 \\
NODE & NN & 1 & 21 & 9.08 & 9 & 0.74 & 0.81 & 0.26 & 0.2 & 138.36 & 117.04 \\
SAINT & NN & 1 & 21 & 9.09 & 8 & 0.73 & 0.86 & 0.31 & 0.24 & 169.54 & 146.16 \\
FTTransformer & NN & 1 & 20 & 9.29 & 9 & 0.76 & 0.8 & 0.31 & 0.21 & 27.67 & 18.4 \\
RandomForest & base & 1 & 21 & 9.5 & 9 & 0.76 & 0.83 & 0.32 & 0.22 & 0.35 & 0.24 \\
LightGBM & GBDT & 1 & 21 & 9.61 & 9 & 0.76 & 0.84 & 0.36 & 0.21 & 0.87 & 0.34 \\
SVM & base & 1 & 20 & 10.34 & 11.5 & 0.69 & 0.76 & 0.26 & 0.19 & 30.4 & 1.67 \\
DANet & NN & 1 & 20 & 11.07 & 11 & 0.73 & 0.79 & 0.32 & 0.23 & 68.82 & 60.15 \\
MLP-rtdl & NN & 1 & 21 & 11.1 & 12 & 0.65 & 0.72 & 0.28 & 0.16 & 14.27 & 7.3 \\
STG & NN & 1 & 21 & 13.19 & 14 & 0.56 & 0.63 & 0.29 & 0.17 & 18.44 & 15.79 \\
DecisionTree & base & 1 & 21 & 13.32 & 15 & 0.59 & 0.68 & 0.35 & 0.25 & 0.03 & 0.01 \\
MLP & NN & 1 & 21 & 13.49 & 15 & 0.57 & 0.57 & 0.29 & 0.18 & 18.39 & 11.2 \\
LinearModel & base & 1 & 21 & 13.76 & 16 & 0.51 & 0.53 & 0.31 & 0.24 & 0.04 & 0.03 \\
TabNet & NN & 1 & 21 & 14.24 & 16 & 0.54 & 0.6 & 0.39 & 0.25 & 34.95 & 29.9 \\
KNN & base & 1 & 21 & 15.32 & 17 & 0.45 & 0.51 & 0.29 & 0.21 & 0.01 & 0.0 \\
VIME & NN & 3 & 21 & 16.73 & 19 & 0.37 & 0.32 & 0.27 & 0.18 & 16.81 & 14.86 \\
\bottomrule
\end{tabular}
}
\end{table}

\begin{table}[t]
\centering
\caption{Performance of algorithms across 98 datasets, where the algorithms include two modified versions of TabPFN.  Columns show the algorithm family (GBDT, NN, baseline, or PFN), rank over all datasets, the average normalized log loss (Mean LL), the std.\ dev.\ of normalized LL across folds (Std.\ LL), and the train time in seconds per 1000 instances. Min/max/mean/median of these quantities are taken over all datasets.}
\label{tab:performance-tabpfn-ll}
\resizebox{\linewidth}{!}{
\begin{tabular}{lcrrrrrrrrrr}
\toprule
{} & {} & \multicolumn{4}{l}{Rank} & \multicolumn{2}{l}{Mean LL} & \multicolumn{2}{l}{Std. LL} & \multicolumn{2}{l}{Time /1000 inst.} \\
Algorithm & Class &               min & max &   mean & med.\ &                           mean & med.\ &                          mean & med.\ &                              mean &  med.\ \\
\midrule
TabPFN-3k & PFN & 1 & 16 & 5.02 & 4 & 0.06 & 0.01 & 0.07 & 0.04 & 0.25 & 0.01 \\
CatBoost & GBDT & 1 & 16 & 5.73 & 5 & 0.05 & 0.02 & 0.08 & 0.06 & 13.89 & 1.66 \\
CatBoost-1k & GBDT & 1 & 18 & 5.93 & 5 & 0.05 & 0.02 & 0.08 & 0.05 & 5.59 & 1.84 \\
TabPFN-1k & PFN & 1 & 17 & 6.21 & 5 & 0.07 & 0.02 & 0.08 & 0.04 & 0.25 & 0.01 \\
XGBoost & GBDT & 1 & 17 & 6.36 & 6 & 0.05 & 0.03 & 0.08 & 0.06 & 0.73 & 0.37 \\
SAINT & NN & 1 & 21 & 8.49 & 8 & 0.12 & 0.06 & 0.09 & 0.08 & 202.59 & 173.23 \\
ResNet & NN & 1 & 18 & 9.67 & 9 & 0.12 & 0.08 & 0.1 & 0.08 & 16.12 & 8.97 \\
SVM & base & 1 & 20 & 9.7 & 10 & 0.15 & 0.06 & 0.08 & 0.05 & 49.83 & 1.2 \\
LightGBM & GBDT & 1 & 21 & 9.76 & 9 & 0.13 & 0.07 & 0.21 & 0.09 & 0.83 & 0.27 \\
DANet & NN & 1 & 20 & 10.08 & 10 & 0.12 & 0.09 & 0.13 & 0.08 & 71.58 & 61.35 \\
FTTransformer & NN & 1 & 19 & 10.85 & 11.5 & 0.14 & 0.09 & 0.11 & 0.09 & 29.58 & 18.48 \\
STG & NN & 1 & 20 & 11.17 & 11 & 0.18 & 0.07 & 0.07 & 0.05 & 18.82 & 15.85 \\
RandomForest & base & 1 & 21 & 12.26 & 13 & 0.19 & 0.13 & 0.22 & 0.07 & 0.29 & 0.22 \\
LinearModel & base & 1 & 21 & 12.28 & 13 & 0.24 & 0.1 & 0.1 & 0.06 & 0.04 & 0.03 \\
MLP-rtdl & NN & 1 & 21 & 13.31 & 14 & 0.28 & 0.18 & 0.17 & 0.12 & 13.75 & 7.96 \\
NODE & NN & 1 & 20 & 13.71 & 14 & 0.23 & 0.19 & 0.04 & 0.03 & 196.82 & 176.16 \\
MLP & NN & 1 & 21 & 14.27 & 15 & 0.29 & 0.21 & 0.15 & 0.11 & 18.29 & 10.95 \\
TabNet & NN & 1 & 21 & 14.55 & 16.5 & 0.4 & 0.25 & 0.38 & 0.19 & 34.62 & 29.69 \\
VIME & NN & 3 & 21 & 16.27 & 18 & 0.4 & 0.37 & 0.09 & 0.07 & 16.92 & 14.64 \\
KNN & base & 1 & 21 & 17.17 & 18 & 0.51 & 0.42 & 0.37 & 0.2 & 0.01 & 0.0 \\
DecisionTree & base & 1 & 21 & 17.46 & 20 & 0.56 & 0.58 & 0.54 & 0.41 & 0.02 & 0.01 \\
\bottomrule
\end{tabular}
}
\end{table}

\begin{table}[t]
\centering
\caption{Performance of algorithms across 98 datasets, where the algorithms include two modified versions of TabPFN.  Columns show the algorithm family (GBDT, NN, baseline, or PFN), rank over all datasets, the average normalized F1 score loss (Mean F1), the std.\ dev.\ of normalized F1 across folds (Std.\ F1), and the train time in seconds per 1000 instances. Min/max/mean/median of these quantities are taken over all datasets.}
\label{tab:performance-tabpfn-f1}
\resizebox{\linewidth}{!}{
\begin{tabular}{lcrrrrrrrrrr}
\toprule
{} & {} & \multicolumn{4}{l}{Rank} & \multicolumn{2}{l}{Mean F1} & \multicolumn{2}{l}{Std. F1} & \multicolumn{2}{l}{Time /1000 inst.} \\
Algorithm & Class &               min & max &   mean & med.\ &                           mean & med.\ &                          mean & med.\ &                              mean &  med.\ \\
\midrule
CatBoost & GBDT & 1 & 19 & 6.43 & 5 & 0.87 & 0.93 & 0.29 & 0.22 & 21.0 & 2.08 \\
TabPFN-3k & PFN & 1 & 20 & 6.46 & 5.5 & 0.83 & 0.93 & 0.25 & 0.19 & 0.25 & 0.01 \\
CatBoost-1k & GBDT & 1 & 21 & 7.09 & 6 & 0.85 & 0.91 & 0.29 & 0.22 & 6.94 & 2.43 \\
XGBoost & GBDT & 1 & 19 & 7.78 & 6 & 0.82 & 0.89 & 0.32 & 0.22 & 0.83 & 0.37 \\
TabPFN-1k & PFN & 1 & 21 & 8.01 & 7 & 0.8 & 0.91 & 0.26 & 0.19 & 0.25 & 0.01 \\
ResNet & NN & 1 & 21 & 8.74 & 8.5 & 0.76 & 0.82 & 0.29 & 0.19 & 16.04 & 9.34 \\
NODE & NN & 1 & 21 & 9.23 & 9 & 0.75 & 0.81 & 0.25 & 0.19 & 140.71 & 117.04 \\
SAINT & NN & 1 & 21 & 9.26 & 9 & 0.73 & 0.85 & 0.3 & 0.23 & 171.14 & 144.37 \\
FTTransformer & NN & 1 & 19 & 9.32 & 9 & 0.76 & 0.82 & 0.3 & 0.19 & 27.94 & 18.4 \\
RandomForest & base & 1 & 21 & 9.46 & 9 & 0.77 & 0.83 & 0.31 & 0.21 & 0.36 & 0.25 \\
LightGBM & GBDT & 1 & 21 & 9.62 & 9 & 0.76 & 0.83 & 0.35 & 0.21 & 0.86 & 0.31 \\
SVM & base & 1 & 20 & 10.3 & 11 & 0.69 & 0.77 & 0.25 & 0.17 & 29.99 & 1.73 \\
MLP-rtdl & NN & 1 & 21 & 10.96 & 12 & 0.66 & 0.73 & 0.27 & 0.16 & 14.29 & 7.3 \\
DANet & NN & 1 & 20 & 11.01 & 11 & 0.74 & 0.81 & 0.31 & 0.22 & 69.54 & 60.2 \\
DecisionTree & base & 1 & 21 & 13.19 & 15 & 0.61 & 0.71 & 0.34 & 0.24 & 0.03 & 0.01 \\
STG & NN & 1 & 21 & 13.24 & 14 & 0.57 & 0.65 & 0.28 & 0.18 & 18.43 & 15.76 \\
MLP & NN & 1 & 21 & 13.69 & 15 & 0.57 & 0.58 & 0.29 & 0.18 & 18.42 & 11.2 \\
LinearModel & base & 1 & 21 & 13.83 & 16 & 0.52 & 0.53 & 0.3 & 0.24 & 0.04 & 0.03 \\
TabNet & NN & 1 & 21 & 14.22 & 16 & 0.56 & 0.6 & 0.38 & 0.26 & 34.82 & 29.16 \\
KNN & base & 1 & 21 & 15.21 & 17 & 0.46 & 0.55 & 0.28 & 0.21 & 0.01 & 0.0 \\
VIME & NN & 3 & 21 & 16.87 & 19 & 0.36 & 0.34 & 0.27 & 0.18 & 17.02 & 14.96 \\
\bottomrule
\end{tabular}
}
\end{table}

